\documentclass{article}

    \usepackage[preprint]{neurips_2019}

\usepackage[utf8]{inputenc} % allow utf-8 input
\usepackage[T1]{fontenc}    % use 8-bit T1 fonts
\usepackage[final]{hyperref}       % hyperlinks
\usepackage{url}            % simple URL typesetting
\usepackage{booktabs}       % professional-quality tables
\usepackage{amsfonts}       % blackboard math symbols
\usepackage{amssymb}
\usepackage{nicefrac}       % compact symbols for 1/2, etc.
\usepackage{microtype}      % microtypography

\usepackage{algorithm,algorithmic}
\usepackage{graphicx}
\usepackage{multirow}
\usepackage{amsmath, bm}
\usepackage{wrapfig}
\usepackage{xfrac}
\usepackage{sidecap}
\usepackage{adjustbox}
\usepackage{caption}
\usepackage{subcaption}
\usepackage{pdflscape}

\usepackage[dvipsnames]{xcolor}

\newcommand{\eps}{\varepsilon}

\newcommand{\mean}{\mathbb{E}}%{{\rm I\kern-.3em E}}
\newcommand{\var}{{\rm I\kern-.3em D}}

\newcommand{\RR}{\mathbb{R}}
\newcommand{\cond}{\,|\,}
\newcommand{\Normal}{\mathcal{N}}

\newcommand{\norm}[1]{\left\lVert#1\right\rVert}
\DeclareMathOperator*{\argmin}{arg\,min}

\newtheorem{proposition}{Proposition}

\title{The Implicit Metropolis-Hastings Algorithm}

\author{%
Kirill Neklyudov \\
Samsung AI Center Moscow \\
Samsung-HSE Laboratory \\
HSE\thanks{National Research University Higher School of Economics}, \; Moscow, Russia \\
\texttt{k.necludov@gmail.com}
  \And
Evgenii Egorov \\
Skoltech\thanks{Skolkovo Institute of Science and Technology}, \; Moscow, Russia \\
\texttt{egorov.evgenyy@ya.ru}
  \And
Dmitry Vetrov \\
Samsung AI Center Moscow \\
Samsung-HSE Laboratory\\
HSE\footnotemark[1], \; Moscow, Russia \\
\texttt{vetrovd@yandex.ru}
}

\begin{document}

\maketitle

\begin{abstract}
% State the problem, your approach and solution, and the main contributions of the paper. Include little if any background and motivation. Be factual but comprehensive. The material in the abstract should not be repeated later word for word in the paper. 

Recent works propose using the discriminator of a GAN to filter out unrealistic samples of the generator.
We generalize these ideas by introducing the implicit Metropolis-Hastings algorithm.
For any implicit probabilistic model and a target distribution represented by a set of samples, implicit Metropolis-Hastings operates by learning a discriminator to estimate the density-ratio and then generating a chain of samples.
Since the approximation of density ratio introduces an error on every step of the chain, it is crucial to analyze the stationary distribution of such chain.
For that purpose, we present a theoretical result stating that the discriminator loss upper bounds the total variation distance between the target distribution and the stationary distribution.
Finally, we validate the proposed algorithm both for independent and Markov proposals on CIFAR-10 and CelebA datasets.
\end{abstract}

\section{Introduction}

% What is the problem?
% Why is it interesting and important?
Learning a generative model from an \textit{empirical} target distribution is one of the key tasks in unsupervised machine learning.
Currently, Generative Adversarial Networks (GANs) \citep{goodfellow2014generative} are among the most successful approaches in building such models.
Unlike conventional sampling techniques, such as Markov Chain Monte-Carlo (MCMC), they operate by learning the \textit{implicit} probabilistic model, which allows for sampling but not for a density evaluation.
Due to the availability of large amounts of empirical data, GANs find many applications in computer vision: image super-resolution \citep{ledig2017photo}, image inpainting \citep{yu2018generative}, and learning representations \citep{donahue2016adversarial}.

% Why is it hard? (E.g., why do naive approaches fail?)
% Why hasn't it been solved before? (Or, what's wrong with previous proposed solutions? How does mine differ?)
Despite the practical success, GANs remain hard for theoretical analysis and do not provide any guarantees on the learned model.
For now, most of the theoretical results assume optimality of the learned discriminator (critic) what never holds in practice \citep{goodfellow2014generative, nowozin2016f, arjovsky2017wasserstein}.
Moreover, there is empirical evidence that GANs do not learn to sample from a target distribution \citep{arora2017gans}.

Recently, the idea of a GAN postprocessing by filtering the generator was proposed in several works.
Under the assumption that the learned discriminator evaluates the exact density-ratio they
filter samples from a generator by rejection sampling \citep{azadi2018discriminator} or by the independent Metropolis-Hastings algorithm \citep{neklyudov2018metropolis, turner2018metropolis}.
Since the assumption of the discriminator optimality never holds in practice, we still cannot be sure that the resulting distribution will be close to the target, we even cannot guarantee that we will improve the output of the generator.

% What are the key components of my approach and results? Also include any specific limitations. 
In this work, we present a theoretical result that justifies the heuristic proposed by \cite{neklyudov2018metropolis, turner2018metropolis} and generalize the proposed algorithm to the case of any implicit probabilistic models --- both independent and Markov.
To do that, we consider some, maybe not optimal, discriminator in the Metropolis-Hastings test, and approach the problem from the MCMC perspective.
Under reasonable assumptions, we derive an upper bound on the total variation distance between the target distribution and the stationary distribution of the produced chain, that can be minimized w.r.t. parameters of the discriminator.

On CIFAR-10 and CelebA datasets, we validate the proposed algorithm using different models as independent proposals: DCGAN \citep{radford2015unsupervised} that is learned in terms of minimax game under a classical GAN approach; Wasserstein GAN with gradient penalty \citep{gulrajani2017improved} that minimizes Wasserstein distance estimated by the critic network; VAE \citep{kingma2013auto} that maximizes the evidence lower bound.
For every proposal, we learn a discriminator from scratch and observe the monotonous improvement of metrics throughout the learning.
Using the generator of WPGAN, we further improve its performance by traversing its latent space via a Markov chain and applying the proposed algorithm.

% Then have a final paragraph or subsection: "Summary of Contributions". It should list the major contributions in bullet form, mentioning in which sections they can be found. This material doubles as an outline of the rest of the paper, saving space and eliminating redundancy. 

We summarize our main contributions as follows.
\begin{itemize}
    \item We propose the implicit Metropolis-Hastings algorithm, that can be seen as an adaptation of the classical Metropolis-Hastings algorithm to the case of an implicit probabilistic model and an empirical target distribution (Section \ref{sec:imh}).
    \item We justify the algorithm proposed by \cite{neklyudov2018metropolis} and \cite{turner2018metropolis}.
    In particular, we demonstrate that learning the discriminator via the binary cross-entropy minimizes an upper bound on the distance between the target distribution and the stationary distribution of the chain (Section \ref{sec:ce}).
    \item We empirically validate the obtained theoretical result on real-world datasets (CIFAR-10, CelebA) (Section \ref{sec:exp_ind}). 
    We also demonstrate empirical gains by applying our algorithm for Markov proposals (Section \ref{sec:exp_markov}).
\end{itemize}

\section{Background}
\label{sec:pre}
\subsection{The Metropolis-Hastings algorithm}
The MH algorithm allows for sampling from an analytic target distribution $p(x)$ by filtering samples from a proposal distribution $q(x\cond y)$ that is also given in the analytic form. 
It operates by sampling a chain of correlated samples that converge in distribution to the target (see Algorithm \ref{alg:MH}).

\begin{minipage}[h]{0.49\textwidth}
\begin{algorithm}[H]
  \caption{The Metropolis-Hastings algorithm}
  \begin{algorithmic}  
    \INPUT{density of target distribution $\hat{p}(x) \propto p(x)$}
    \INPUT{proposal distribution $q(x\cond y)$}
    \STATE $y \gets$ random init
    \FOR{$i = 0\ldots n$}
        \STATE sample proposal point $x \sim q(x\cond y)$
        \STATE $P = \min\{1,\frac{\hat{p}(x)q(y\cond x)}{\hat{p}(y)q(x\cond y)} \}$
        \STATE $ x_{i} =
            \begin{cases} 
            x , \text{ with probability } P\\
            y , \text{ with probability } (1-P)
            \end{cases}$
        \STATE $y \gets x_{i}$
    \ENDFOR
    \OUTPUT{ $\{x_0,\ldots, x_n\}$}
  \end{algorithmic} 
  \label{alg:MH}
\end{algorithm}
\end{minipage}
\hfill
\begin{minipage}[h]{0.49\textwidth}
\begin{algorithm}[H]
  \caption{Metropolis-Hastings GAN}
  \begin{algorithmic}  
    \INPUT{target dataset $\mathcal{D}$} 
    \INPUT{learned generator $q(x)$, discriminator $d(\cdot)$}
    \STATE $y \sim \mathcal{D}$ initialize from the dataset
    \FOR{$i = 0\ldots n$}
        \STATE sample proposal point $x \sim q(x)$
        \STATE $P = \min\{1,\frac{d(x)(1-d(y))}{(1-d(x))d(y)} \}$
        \STATE $ x_{i} =
            \begin{cases} 
            x , \text{ with probability } P \\
            y , \text{ with probability } (1-P)
            \end{cases}$
        \STATE $y \gets x_{i}$
    \ENDFOR
    \OUTPUT{ $\{x_0,\ldots, x_n\}$}
  \end{algorithmic} 
  \label{alg:MH_disc}
\end{algorithm}
\end{minipage}

If we take a proposal distribution that is not conditioned on the previous point, we will obtain the \textbf{independent} MH algorithm.
It operates in the same way, but samples all of the proposal points independently $q(x\cond y) = q(x)$.

\subsection{Metropolis-Hastings GAN}
Recent works \citep{neklyudov2018metropolis, turner2018metropolis} propose to treat the generator of a GAN as an independent proposal distribution $q(x)$ and perform an approximate Metropolis-Hastings test via the discriminator.
Authors motivate this approximation by the fact that the optimal discriminator evaluates the true density-ratio
\begin{align}
    d^*(x) = \frac{p(x)}{p(x) + q(x)} = \argmin_d \bigg[-\mean_{x\sim p(x)} \log d(x) -\mean_{x \sim q(x)} \log (1-d(x)) \bigg].
    \label{eq:disc_opt}
\end{align}
Substituting the optimal discriminator in the acceptance test, one can obtain the Metropolis-Hastings correction of a GAN, that is described in Algorithm \ref{alg:MH_disc}.

In contrast to the previous works, we take the non-optimality of the discriminator as given and analyze the stationary distribution of the resulting chain for both independent and Markov proposals.
In Section \ref{sec:imh}, we formulate the implicit Metropolis-Hastings algorithm and derive an upper bound on the total variation distance between the target distribution and the stationary distribution of the chain.
Then, in Appendix \ref{app:ce}, we justify Algorithm \ref{alg:MH_disc} by relating the obtained upper bound with the binary cross-entropy.

\section{The Implicit Metropolis-Hastings Algorithm}
\label{sec:imh}

\begin{wrapfigure}{R}{0.5\textwidth}
    \begin{minipage}{0.5\textwidth}
    \vskip -0.3in
      \begin{algorithm}[H]
      \caption{\\The implicit Metropolis-Hastings algorithm}
      \begin{algorithmic}  
        \INPUT{target dataset $\mathcal{D}$} 
        \INPUT{implicit model $q(x\cond y)$}
        \INPUT{learned discriminator $d(\cdot,\cdot)$}
        \STATE $y \sim \mathcal{D}$ initialize from the dataset
        \FOR{$i = 0\ldots n$}
            \STATE sample proposal point $x \sim q(x \cond y)$
            \STATE $P = \min\{1,\frac{d(x,y)}{d(y,x)}\}$
            \STATE $ x_{i} =
                \begin{cases} 
                x , \text{ with probability } P\\
                y , \text{ with probability } (1-P)
                \end{cases}$
            \STATE $y \gets x_{i}$
        \ENDFOR
        \OUTPUT{ $\{x_0,\ldots, x_n\}$}
      \end{algorithmic} 
      \label{alg:IMH}
    \end{algorithm}
    \end{minipage}
\end{wrapfigure}

In this section, we describe the implicit Metropolis-Hastings algorithm and present a theoretical analysis of its stationary distribution.

The Implicit Metropolis-Hastings algorithm is aimed to sample from an empirical target distribution $p(x),\; x \in \RR^D$, while being able to sample from an implicit proposal distribution $q(x\cond y)$.
Given a discriminator $d(x,y)$, it generates a chain of samples as described in Algorithm \ref{alg:IMH}.

We build our reasoning by first assuming that the chain is generated using some discriminator and then successively introducing conditions on the discriminator and upper bounding the distance between the chain and the target.
Finally, we come up with an upper bound that can be minimized w.r.t. parameters of the discriminator.
Here we consider the case of an implicit Markov proposal, but all of the derivations also hold for independent proposals.

The transition kernel of the implicit Metropolis-Hastings algorithm is
\begin{align}
    t(x\cond y) =  q(x \cond y) \min\bigg\{1, \frac{d(x,y)}{d(y,x)}\bigg\} + \delta(x-y) \int dx' q(x' \cond y)\bigg(1 - \min\bigg\{1, \frac{d(x',y)}{d(y,x')}\bigg\}\bigg).
\label{eq:imh_t}
\end{align}
First of all, we need to ensure that the Markov chain defined by the transition kernel $t(x\cond y)$ converges to some stationary distribution $t_\infty(x)$.
In order to do that, we require the proposal distribution $q(x\cond y)$ and the discriminator $d(x,y)$ to be \textit{continuous} and \textit{positive} on $\RR^D\times\RR^D$.
In Appendix \ref{app:prop}, we show that these requirements guarantee the following properties of the transition kernel $t$:
\begin{itemize}
    \item the kernel $t$ defines a correct conditional distribution;
    \item the Markov chain defined by $t$ is \textit{irreducible};
    \item the Markov chain defined by $t$ is \textit{aperiodic}.
\end{itemize}
These properties imply convergence of the Markov chain defined by $t$ to some stationary distribution $t_\infty$ \citep{roberts2004general}.

Further, we want the stationary distribution $t_\infty$ of our Markov chain to be as close as possible to the target distribution $p$.
To measure the closeness of distributions, we consider a standard metric for analysis in MCMC --- the \textit{total variation distance}
\begin{equation}
    \norm{t_\infty - p}_{TV} = \frac{1}{2} \int |t_\infty(x) - p(x)|dx.
\end{equation}
We assume the proposal $q(x\cond y)$ to be given, but different $d(x,y)$ may lead to different $t_\infty$.
That is why we want to derive an upper bound on the distance $\norm{t_\infty - p}_{TV}$ and minimize it w.r.t. parameters of the discriminator $d(x,y)$.
We derive this upper bound in three steps in the following subsections.

\subsection{Fast convergence}
In practice, estimation of the stationary distribution $t_\infty$ by running a chain is impossible.
Nevertheless, if we know that the chain converges fast enough, we can upper bound the distance $\norm{t_\infty - p}_{TV}$ using the distance $\norm{t_1 - p}_{TV}$, where $t_1$ is the one-step distribution $t_1(x) = \int t(x\cond y) t_0(y) dy$, and $t_0$ is some initial distribution of the chain.

To guarantee fast convergence of a chain, we propose to use the \textit{minorization condition} \citep{roberts2004general}. For a transition kernel $t(x\cond y)$, it requires that exists such $\eps > 0$ and a distribution $\nu$ that the following condition is satisfied
\begin{equation}
    t(x\cond y) > \eps \nu(x) \;\; \forall (x,y) \in \mathbb{R}^D\times\mathbb{R}^D.
\end{equation}
When a transition kernel satisfies the minorization condition, the Markov chain converges "fast" to the stationary distribution.
We formalize this statement in the following Proposition.
\begin{proposition}
Consider a transition kernel $t(x\cond y)$ that satisfies the minorization condition $t(x\cond y) > \eps \nu(x)$ for some $\eps > 0$, and distribution $\nu$.
Then the distance between two consequent steps decreases as:
\begin{equation}
    \norm{t_{n+2} - t_{n+1}}_{TV} \leq (1-\eps)\norm{t_{n+1} - t_{n}}_{TV},
\end{equation}
where distribution $t_{k+1}(x) = \int t(x\cond y) t_k(y) dy$.
\label{th:uniform}
\end{proposition}
This result could be considered as a corollary of Theorem 8 in \cite{roberts2004general}.
For consistency, we provide an independent proof of Proposition \ref{th:uniform} in Appendix \ref{app:uniform}.

To guarantee minorization condition of our transition kernel $t(x\cond y)$, we require the proposal $q(x\cond y)$ to satisfy minorization condition with some constant $\eps$ and distribution $\nu$ (note that for an independent proposal, the minorization condition holds automatically with $\eps=1$).
Also, we limit the range of the discriminator as $d(x,y) \in [b,1] \; \forall x,y$, where $b$ is some positive constant that can be treated as a hyperparameter of the algorithm.
These requirements imply
\begin{equation}
    t(x\cond y) \geq b q(x \cond y) > b\eps \nu(x).
\label{eq:min_cond}
\end{equation}
Using Proposition \ref{th:uniform} and minorization condition \eqref{eq:min_cond} for $t$, we can upper bound the TV-distance between an initial distribution $t_0$ and the stationary distribution $t_\infty$ of implicit Metropolis-Hastings.
\begin{align}
    \norm{t_{\infty} - t_{0}}_{TV} \leq \sum_{i=0}^\infty \norm{t_{i+1} - t_{i}}_{TV} \leq \sum_{i=0}^\infty (1-b\eps)^i\norm{t_{1} - t_{0}}_{TV} = \frac{1}{b\eps}\norm{t_{1} - t_{0}}_{TV}
\label{eq:stat_bound}
\end{align}
Taking the target distribution $p(x)$ as the initial distribution $t_0(x)$ of our chain $t(x\cond y)$, we reduce the problem of estimation of the distance $\norm{t_{\infty} - p}_{TV}$ to the problem of estimation of the distance $\norm{t_{1} - p}_{TV}$:
\begin{equation}
    \norm{t_{\infty} - p}_{TV} \leq \frac{1}{b\eps}\norm{t_{1} - p}_{TV} = \frac{1}{b\eps} \cdot\frac{1}{2} \int dx \bigg|\int t(x\cond y)p(y) dy - p(x)\bigg|.
    \label{eq:pt_bound}
\end{equation}
However, the estimation of this distance raises two issues.
Firstly, we need to get rid of the inner integral $\int t(x\cond y)p(y) dy$.
Secondly, we need to bypass the evaluation of densities $t(x\cond y)$ and $p(x)$.
We address these issues in the following subsections.

\subsection{Dealing with the integral inside of the nonlinearity}
For now, assume that we have access to the densities $t(x\cond y)$ and $p(x)$.
However, evaluation of the density $t_1(x)$ is an infeasible problem in most cases.
To estimate $t_1(x)$, one would like to resort to the Monte-Carlo estimation:
\begin{equation}
    t_1(x) = \int t(x\cond y) p(y) dy = \mean_{y\sim p(y)} t(x\cond y).
\end{equation}
However, straightforward estimation of $t_1(x)$ results in a biased estimation of $\norm{t_1 - p}_{TV}$, since the expectation is inside of a nonlinear function.
To overcome this problem, we upper bound this distance in the following proposition.
\begin{proposition}
For the kernel $t(x\cond y)$ of the implicit Metropolis-Hastings algorithm, the distance between initial distribution $p(x)$ and the distribution $t_1(x)$ has the following upper bound
\begin{align}
    \norm{t_1 - p}_{TV} \leq 2\norm{q(y\cond x)p(x) - q(x \cond y)p(y) \frac{d(x,y)}{d(y,x)}}_{TV},
\label{eq:tv_bound}
\end{align}
where the TV-distance on the right side is evaluated in the joint space $(x,y) \in \mathbb{R}^D\times\mathbb{R}^D$.
\label{th:tv}
\end{proposition}
For the proof of this proposition, see Appendix \ref{app:step}.
Note that the obtained upper bound no longer requires evaluation of an integral inside of a nonlinear function.
Moreover, the right side of \eqref{eq:tv_bound} has a reasonable motivation since it is an averaged $l_1$ error of the density ratio estimation.
\begin{equation}
    \norm{q(y\cond x)p(x) - q(x \cond y)p(y) \frac{d(x,y)}{d(y,x)}}_{TV} = 
    \frac{1}{2}\int p(y)q(x\cond y)\bigg| \frac{q(y\cond x)p(x)}{q(x \cond y)p(y)}-\frac{d(x,y)}{d(y,x)}\bigg| dx dy
\end{equation}
In this formulation, we see that we still could achieve zero value of $\norm{t_1 - p}_{TV}$ if we could take such discriminator that estimates the desired density ratio $\frac{d(x,y)}{d(y,x)} = \frac{q(y\cond x)p(x)}{q(x \cond y)p(y)}$.

\subsection{Dealing with the evaluation of densities}
For an estimation of the right side of \eqref{eq:tv_bound}, we still need densities $p(x)$ and $q(x\cond y)$.
To overcome this issue, we propose to upper bound the obtained TV distance via KL-divergence.
Then we show that obtained KL divergence decomposes into two terms: the first term requires evaluation of densities but does not depend on the discriminator $d(x,y)$, and the second term can be estimated only by evaluation of $d(x,y)$ on samples from $p(x)$ and $q(x\cond y)$.

To upper bound the TV-distance $\norm{\alpha - \beta}_{TV}$ via KL-divergence $\mathrm{KL}(\alpha\Vert\beta)$ one can use well-known Pinsker's inequality:
\begin{equation}
    2\norm{\alpha - \beta}_{TV}^2 \leq \mathrm{KL}(\alpha\Vert \beta).
\end{equation}
However, Pinsker's inequality assumes that both $\alpha$ and $\beta$ are distributions, while it is not always true for function $q(x \cond y)p(y) \frac{d(x,y)}{d(y,x)}$ in \eqref{eq:tv_bound}.
In the following proposition, we extend Pinsker's inequality to the case when one of the functions is not normalized.

\begin{proposition}
\label{th:pinsker}
For a distribution $\alpha(x)$ and some positive function $f(x) > 0 \; \forall x$ the following inequality holds:
\begin{equation}
    \norm{\alpha - f}_{TV}^2 \leq \bigg(\frac{2C_f + 1}{6}\bigg) (\widehat{\mathrm{KL}}(\alpha\Vert f) + C_f - 1),
\label{eq:pinsker}
\end{equation}
where $C_f$ is the normalization constant of function $f$: $C_f = \int f(x) dx$,
and $\widehat{\mathrm{KL}}(\alpha\Vert f)$ is the formal evaluation of the KL divergence
\begin{equation}
    \widehat{\mathrm{KL}}(\alpha\Vert f) = \int \alpha(x) \log\frac{\alpha(x)}{f(x)}dx.
\end{equation}
\end{proposition}
The proof of the proposition is in Appendix \ref{app:pinsker}.

Now we use this proposition to upper bound the right side of \eqref{eq:tv_bound}:
\begin{gather}
    \norm{q(y\cond x)p(x) - q(x \cond y)p(y) \frac{d(x,y)}{d(y,x)}}_{TV}^2 \leq \label{eq:pt1_bound}\\ \leq\bigg(\frac{2C + 1}{6}\bigg) \bigg(\widehat{\mathrm{KL}}\bigg(q(y\cond x)p(x)\bigg\Vert q(x \cond y)p(y) \frac{d(x,y)}{d(y,x)}\bigg) + C - 1\bigg). \nonumber
\end{gather}
Here $C$ is the normalization constant of $q(x \cond y)p(y) \frac{d(x,y)}{d(y,x)}$.
For the multiplicative term $(2C+1)/6$, we upper bound $C$ as
\begin{equation}
    C = \int q(x \cond y)p(y) \frac{d(x,y)}{d(y,x)} dxdy \leq \int q(x \cond y)p(y) \frac{1}{b} dxdy = \frac{1}{b},
\label{eq:c_bound}
\end{equation}
since we limit the range of the discriminator as $d(x,y) \in [b,1] \;\; \forall x,y$.

Summing up the results \eqref{eq:pt_bound}, \eqref{eq:tv_bound}, \eqref{eq:pt1_bound}, \eqref{eq:c_bound}, we obtain the final upper bound as follows.
\begin{gather}
    \norm{t_{\infty} - p}_{TV}^2 \leq \frac{1}{b^2\eps^2}\norm{t_{1} - p}_{TV}^2  \leq 
    \frac{4}{b^2\eps^2}\norm{q(y\cond x)p(x) - q(x \cond y)p(y) \frac{d(x,y)}{d(y,x)}}_{TV}^2  \leq \label{eq:final_bound}\\
    \leq \bigg(\frac{4+2b}{3\eps^2 b^3}\bigg)
    \bigg( \underbrace{\text{\Large $\mean$}_{\text{\footnotesize $\begin{matrix}
    x \sim p(x) \\ 
    y \sim q(y\cond x)
    \end{matrix}$}} \bigg[ \log\frac{d(y,x)}{d(x,y)} + \frac{d(y,x)}{d(x,y)} \bigg]}_{\text{loss for the discriminator}} - 1 +  \mathrm{KL}\bigg(q(y\cond x)p(x)\bigg\Vert q(x \cond y)p(y)\bigg)\bigg) \nonumber
\end{gather}

Minimization of the resulting upper bound w.r.t. the discriminator $d(x,y)$ is equivalent to the following optimization problem:
\begin{equation}
    \min_d \text{\Large $\mean$}_{\text{\footnotesize $\begin{matrix}
    x \sim p(x) \\ 
    y \sim q(y\cond x)
    \end{matrix}$}} \bigg[ \log\frac{d(y,x)}{d(x,y)} + \frac{d(y,x)}{d(x,y)} \bigg].
\label{eq:main_loss}
\end{equation}
Thus, we derive the loss function that we can unbiasedly estimate and minimize w.r.t. parameters of $d(x,y)$.
We analyze the optimal solution in the following subsection.

\subsection{The optimal discriminator}
By taking the derivative of objective \eqref{eq:main_loss}, we show (see Appendix \ref{app:dre}) that the optimal discriminator $d^*$ must satisfy 
\begin{equation}
    \frac{d^*(x,y)}{d^*(y,x)} = \frac{q(y\cond x)p(x)}{q(x \cond y)p(y)}.
    \label{eq:optimal_disc}
\end{equation}
When the loss function \eqref{eq:main_loss} achieves its minimum, it becomes
\begin{align}
    \text{\Large $\mean$}_{\text{\footnotesize $\begin{matrix}
    x \sim p(x) \\ 
    y \sim q(y\cond x)
    \end{matrix}$}} \bigg[ \log\frac{q(x \cond y)p(y)}{q(y\cond x)p(x)} + \frac{q(x \cond y)p(y)}{q(y\cond x)p(x)} \bigg] 
    = - \mathrm{KL}\bigg(q(y\cond x)p(x)\bigg\Vert q(x \cond y)p(y)\bigg) + 1
\end{align}
Substituting this equation into \eqref{eq:final_bound}, we achieve $\norm{t_{\infty} - p}_{TV} = 0$.
However, since we limit the range of the discriminator $d(x,y) \in [b, 1]$, the optimal solution could be achieved only when the density-ratio lies in the following range:
\begin{equation}
    \forall x,y \;\;\;\; \frac{q(y\cond x)p(x)}{q(x \cond y)p(y)} \in [b,b^{-1}].
\end{equation}
Therefore, $b$ should be chosen small enough that range $[b,b^{-1}]$ includes all the possible values of density-ratio.
Such $b > 0$ exists if the support of the target distribution is \textit{compact}.
Indeed, if we have positive $p(x)$ and $q(x\cond y)$ on compact support, we can find a minimum of the density-ratio and set $b$ to that minimum.
Moreover, taking a positive $q(x\cond y)$ on a compact support yields minorization condition for the $q(x\cond y)$.

If the support of target distribution is not compact, we may resort to the approximation of the target distribution on some smaller compact support that contains say $99.9\%$ of the whole mass of target distribution.
In practice, many problems of generative modeling are defined on compact support, e.g. the distribution of images lies in finite support since we represent an image by pixels values.
% In practice in many problems of generative modeling (e.g. generation of images) the target distribution is defined in a finite support hence we may guarantee the existence of such $b$ that the equation \eqref{eq:optimal_disc} holds.

\subsection{Relation to the cross-entropy}
\label{sec:ce}
It is possible to upper bound the loss \eqref{eq:main_loss} by the binary cross-entropy.
For a Markov proposal, it is
\begin{equation}
    \text{\Large $\mean$}_{\text{\footnotesize $\begin{matrix}
    x \sim p(x) \\ 
    y \sim q(y\cond x)
    \end{matrix}$}} \bigg[ \log\frac{d(y,x)}{d(x,y)} + \frac{d(y,x)}{d(x,y)} \bigg] \leq \text{\Large $\mean$}_{\text{\footnotesize $\begin{matrix}
    x \sim p(x) \\ 
    y \sim q(y\cond x)
    \end{matrix}$}} \bigg[ -\log d(x,y) - \log (1-d(y,x)) + \frac{1}{b} \bigg].
\end{equation}
In the case of an independent proposal, we factorize the discriminator as $d(x,y) = d(x)(1-d(y))$ and obtain the following inequality (see Appendix \ref{app:ce}).
\begin{equation}
    \text{\Large $\mean$}_{\text{\footnotesize $\begin{matrix}
    x \sim p(x) \\ 
    y \sim q(y\cond x)
    \end{matrix}$}} \bigg[ \log\frac{d(y,x)}{d(x,y)} + \frac{d(y,x)}{d(x,y)} \bigg] \leq -\mean_{x\sim p(x)} \log d(x) -\mean_{y\sim q(y)} \log (1-d(y)) + \frac{1}{b}
\end{equation}
Thus, learning a discriminator via the binary cross-entropy, we also minimize the distance $\norm{t_{\infty} - p}_{TV}$.
This fact justifies Algorithm \ref{alg:MH_disc}.

\begin{table}[t]
\def\arraystretch{2.5}
\caption{Different losses for a density-ratio estimation.}
\label{tab:losses}
\begin{center}
\vskip -0.15in
\begin{small}
% \begin{sc}
\resizebox{\textwidth}{!}{
\begin{tabular}{lll}
% \toprule
\multicolumn{1}{c}{\bf Probosal} & \multicolumn{1}{c}{\bf Name} & \multicolumn{1}{c}{\bf Loss}\\
\hline
\multirow{2}{*}{Markov} & Upper bound (UB) & $\displaystyle\int dxdy~p(x)q(y\cond x)\bigg[ \log\frac{d(y,x)}{d(x,y)} + \frac{d(y,x)}{d(x,y)}\bigg]$  \\
    & Markov cross-entropy (MCE) & $\displaystyle\int dxdy~p(x)q(y\cond x)[-\log d(x,y) -\log (1-d(y,x))]$  \\
    \hline
Independent & Conventional cross-entropy (CCE) & $\displaystyle\int dxdy~p(x)q(y)[-\log d(x)(1-d(y))]$  \\
% \bottomrule
\end{tabular}}
% \end{sc}
\end{small}
\vskip -0.2in
\end{center}
\end{table}

\section{Experiments}
\label{sec:exp}
We present an empirical evaluation of the proposed algorithm and theory for both independent and Markov proposals.
In both cases sampling via the implicit MH algorithm is better than the straightforward sampling from a generator.
For independent proposals, we validate our theoretical result by demonstrating monotonous improvements of the sampling procedure throughout the learning of the discriminator.
Further, the implicit MH algorithm with a Markov proposal compares favorably against Algorithm \ref{alg:MH_disc} proposed by \citep{neklyudov2018metropolis, turner2018metropolis}.
Code reproducing all experiments is available online\footnote{\url{https://github.com/necludov/implicit-MH}}.

Since one can evaluate the total variation distance only when explicit densities are given, we show its monotonous fall only for a synthetic example (Appendix \ref{app:synth}).
For complex empirical distributions, we consider the problem of sampling from the space of images (CIFAR-10 and CelebA datasets) and resort to the conventional metrics for the performance evaluation: the Inception Score (IS) \citep{salimans2016improved} and Frechet Inception Distance (FID) \citep{heusel2017gans}. 
Note that these metrics rely heavily on the implementation of Inception network \citep{barratt2018note}; therefore, for all experiments, we use PyTorch version of the Inception V3 network \citep{paszke2017automatic}.

\subsection{Independent proposals}
\label{sec:exp_ind}
Since we propose to use the implicit MH algorithm for any implicit sampler, we consider three models that are learned under completely different approaches: Wasserstein GAN with gradient penalty (WPGAN) \citep{gulrajani2017improved}, Deep Convolutional GAN (DCGAN) \citep{radford2015unsupervised}, Variational Auto-Encoder (VAE) \citep{kingma2013auto}.
To run the MH algorithm, we treat these models as independent proposals and learn the discriminator for acceptance test from scratch.

Our theoretical result says that the total variation distance between the stationary distribution and the target can be upper bounded by different losses (see Table \ref{tab:losses}).
Note, that we also can learn a discriminator by UB and MCE for independent proposals; however, in practice, we found that CCE performs slightly better.
In Figure \ref{fig:iters}, we demonstrate that the minimization of CCE leads to better IS and FID throughout the learning of a discriminator (see plots for DCGAN in Appendix \ref{app:mono_plots}).

\begin{figure}[h]
    \centering
    \includegraphics[width=0.24\textwidth]{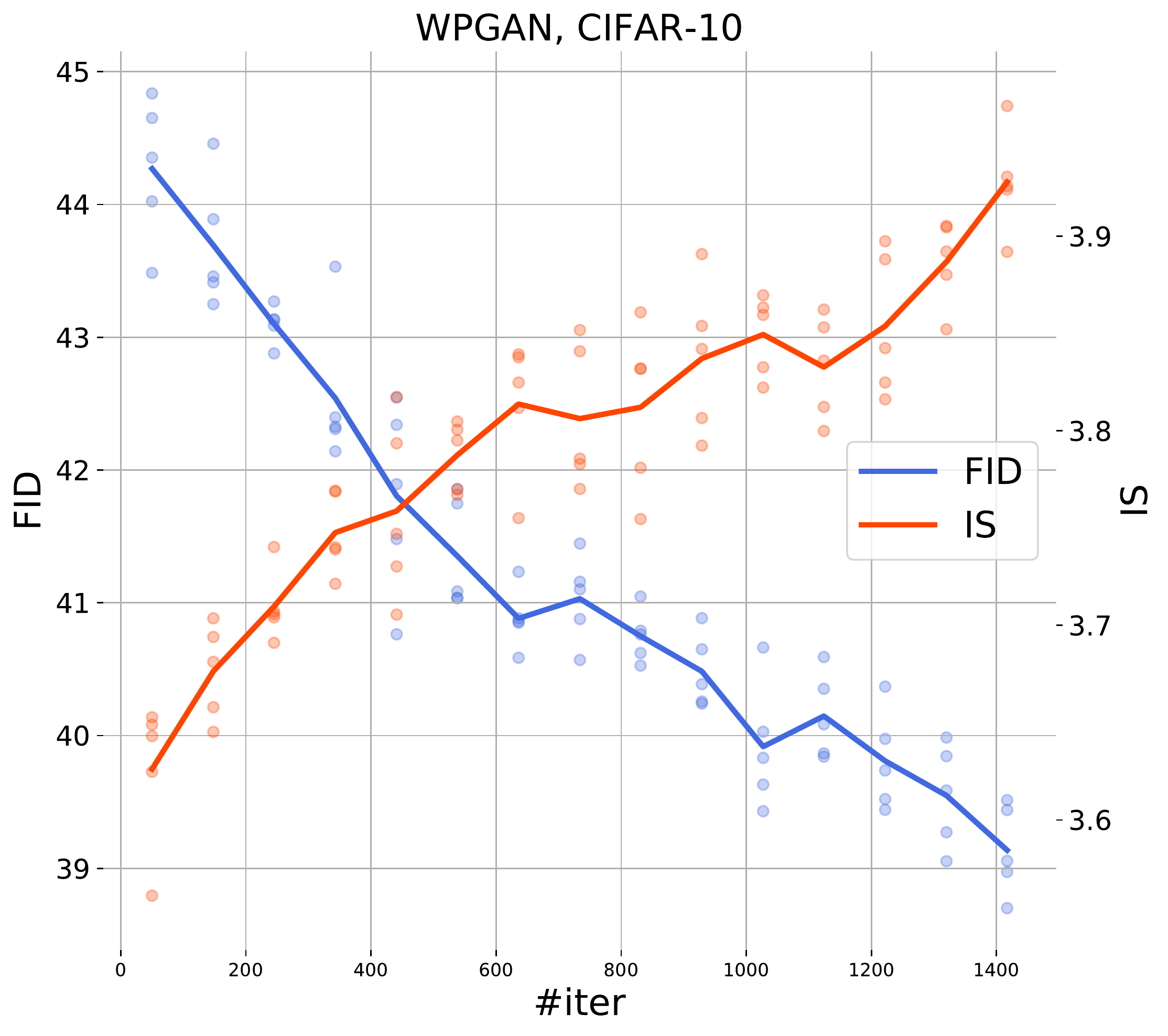}
    \includegraphics[width=0.24\textwidth]{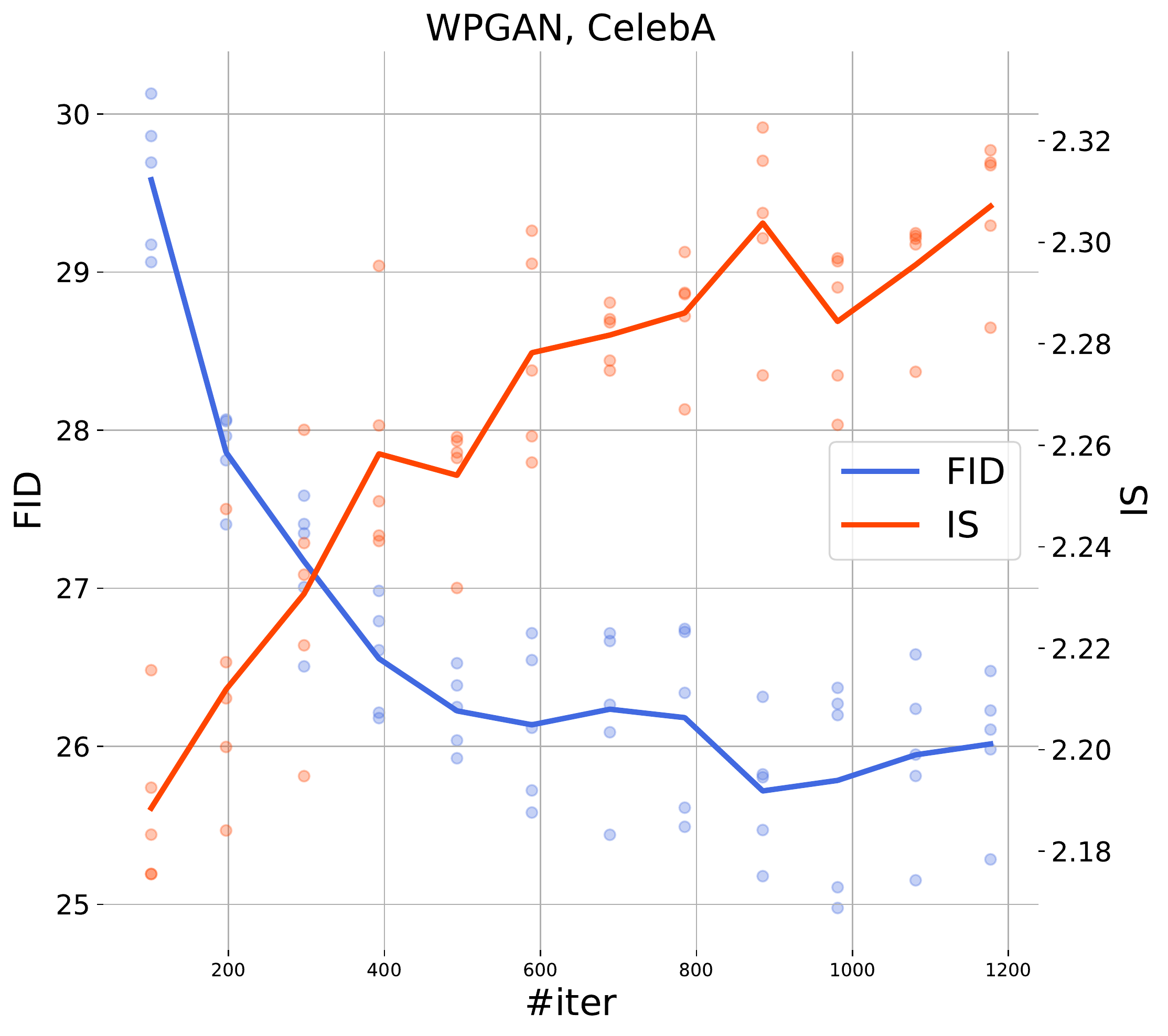}
    \includegraphics[width=0.24\textwidth]{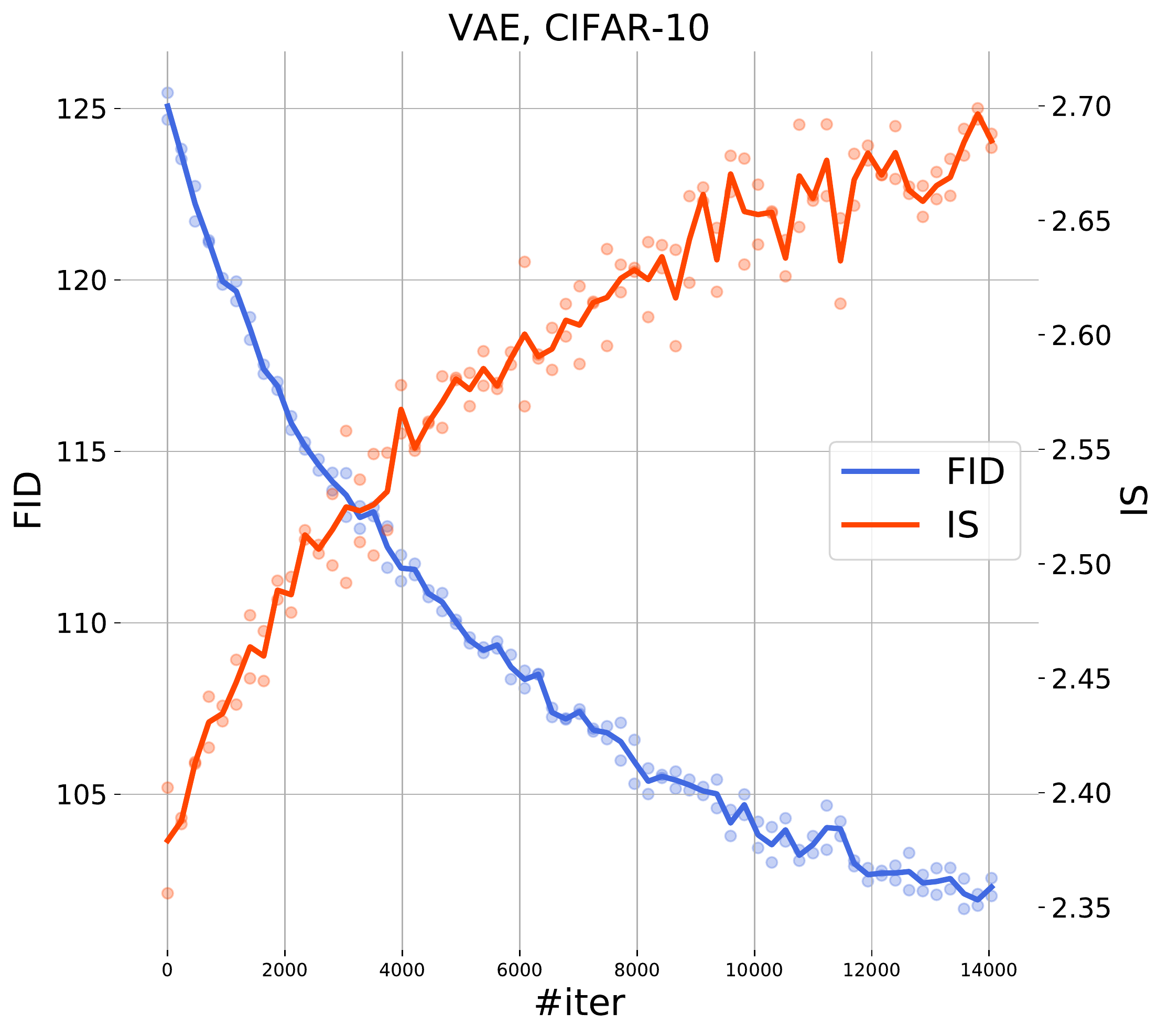}
    \includegraphics[width=0.24\textwidth]{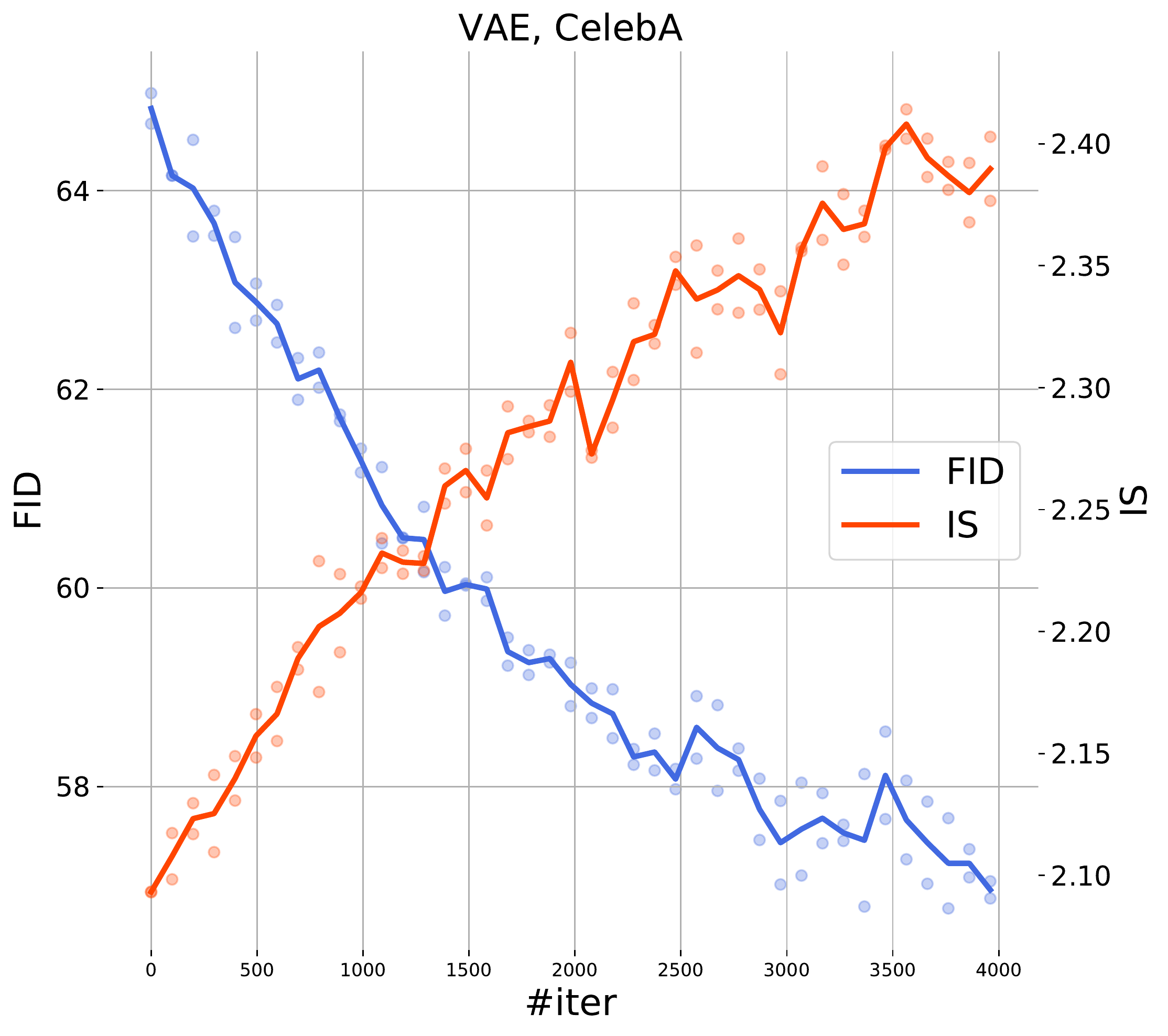}
    \caption{Monotonous improvements in terms of FID and IS for the learning of discriminator by CCE.
    During iterations, we evaluate metrics $5$ times (scatter) and then average them (solid lines).
    For a single metric evaluation, we use 10k samples. 
    Higher values of IS and lower values of FID are better.
    Performance for the original generator corresponds to $0$th iteration of a discriminator.}
    \label{fig:iters}
\end{figure}

% \begin{table}[t]
% \caption{Comparison of sampling using the MH algorithm and using the generator for different models. 
% Low FID and high IS are better.
% For a single evaluation of metrics on CIFAR-10 and CelebA datasets, we use 10k samples, and on ImageNet, we use 50k samples. 
% Then we average all the values across 5 independent runs. 
% See the description of models in the text.}
% \label{tab:gans}
% \vskip -0.15in
% \begin{center}
% \begin{small}
% % \begin{sc}
% \resizebox{\textwidth}{!}{
% \begin{tabular}{lccccccc}
%     \toprule
%     \multicolumn{2}{c}{} & \multicolumn{2}{c}{\bf WPGAN} & \multicolumn{2}{c}{\bf DCGAN} & \multicolumn{2}{c}{\bf VAE} \\
%     \midrule
%     & & Generator & IMH & Generator & IMH & Generator & IMH  \\
%     \midrule
%     \multirow{2}{*}{CIFAR-10} & FID & - & - & - & - & - & -\\
%      & IS & - & - & - & - & - & -\\
%     \midrule 
%     \multirow{2}{*}{CelebA} & FID & - & - & - & - & - & -\\
%      & IS & - & - & - & - & - & -\\
%     \bottomrule
% \end{tabular}}
% % \end{sc}
% \end{small}
% \end{center}
% \end{table}

\subsection{Markov proposals}
\label{sec:exp_markov}
\begin{figure}[h]
    \centering
    \includegraphics[width=\textwidth]{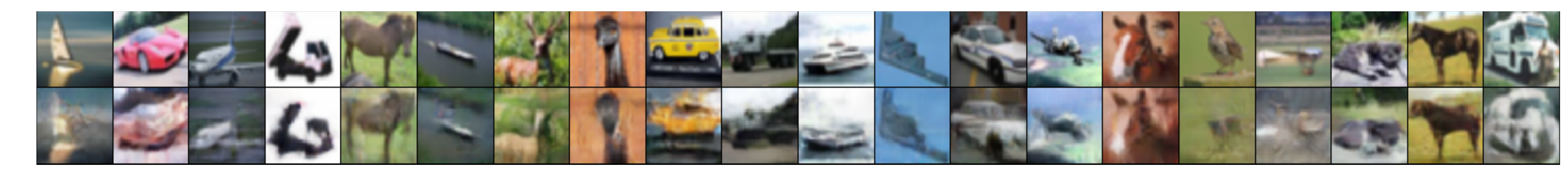}
    \caption{Samples from CIFAR-10 (top line) and their reconstructions (bottom line)}
    \label{fig:rec}
\end{figure}

To simulate Markov proposals we take the same WPGAN as in the independent case and traverse its latent space by a Markov chain.
Taking the latent vector $z_y$ for the previous image $y$, we sample the next vector $z_x$ via HMC and obtain the next image $x = g(z_x)$ by the generator $g(\cdot)$, thus simulating a Markov proposal $q(x\cond y)$.
Sampling via HMC from the Gaussian is equivalent to the interpolation between the previous accepted point $z_y$ and the random vector $v$:
\begin{equation}
    z_x = \cos(t) z_y + \sin(t) v, \;\;\; v \sim \Normal(0,I).
\end{equation}
In our experiments, we take $t=\pi/3$.
For loss estimation, we condition samples from the proposal on samples from the dataset $x \sim q(x\cond y), y\sim p(y)$. 
However, to sample an image $x \sim q(x\cond y)$ we need to know the latent vector $z_y$ for an image $y$ from the dataset.
We find such vectors by optimization in the latent space, minimizing the $l_2$ reconstruction error (reconstructions are in Fig. \ref{fig:rec}).

To filter a Markov proposal, we need to learn a pairwise discriminator, as suggested in Section \ref{sec:imh}.
For this purpose, we take the same architecture of the discriminator as in the independent case and put the difference of its logits $\text{net}(\cdot)$ into the sigmoid.
\begin{equation}
    d(x,y) = \frac{1}{1+\exp(\text{net}(y)-\text{net}(x))}
\end{equation}
Then we learn this discriminator by minimization of UB and MCE (see Table 1).

\begin{figure}[h]
    \centering
    \includegraphics[width=0.24\textwidth]{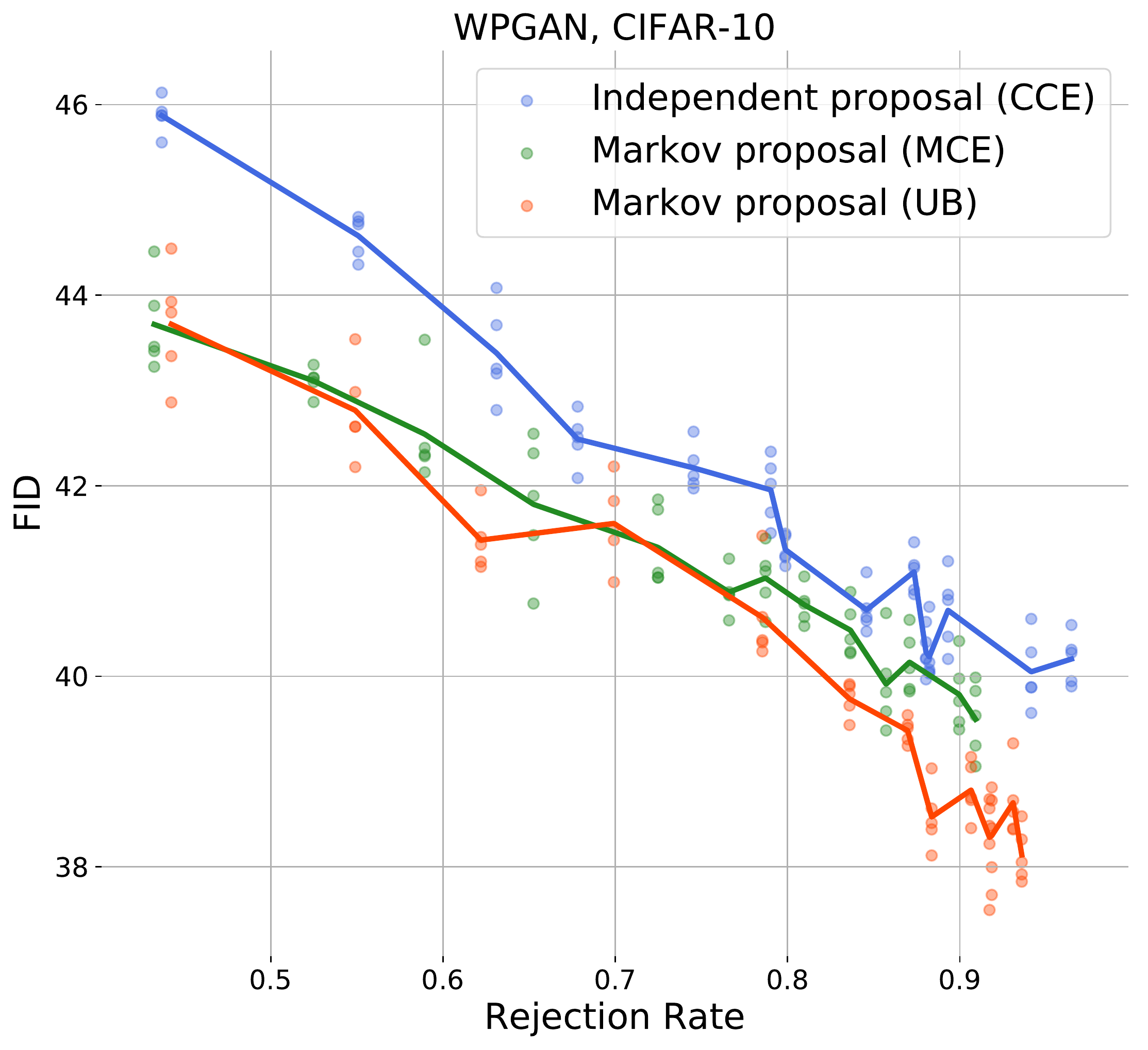}
    \includegraphics[width=0.24\textwidth]{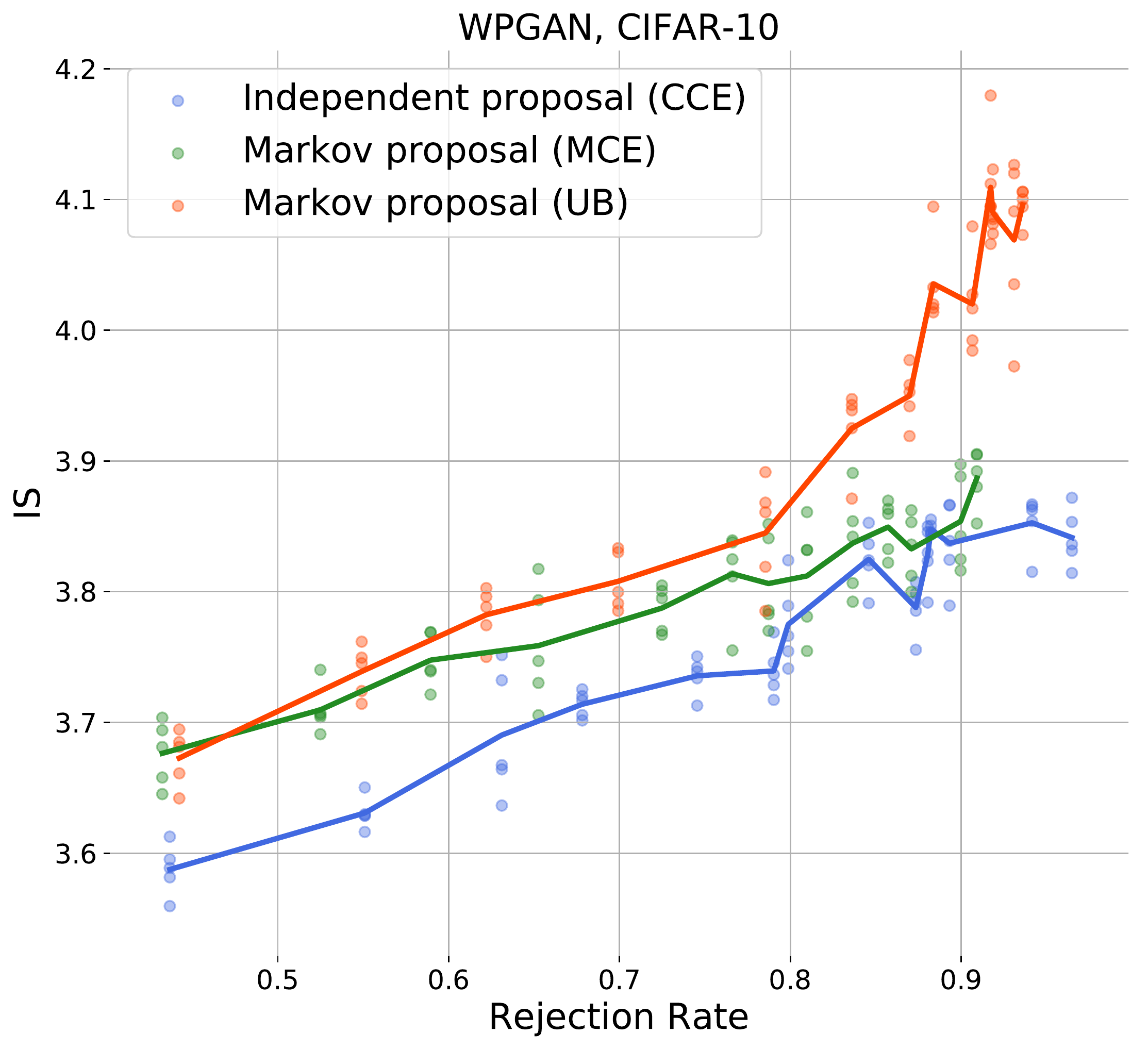}
    \includegraphics[width=0.24\textwidth]{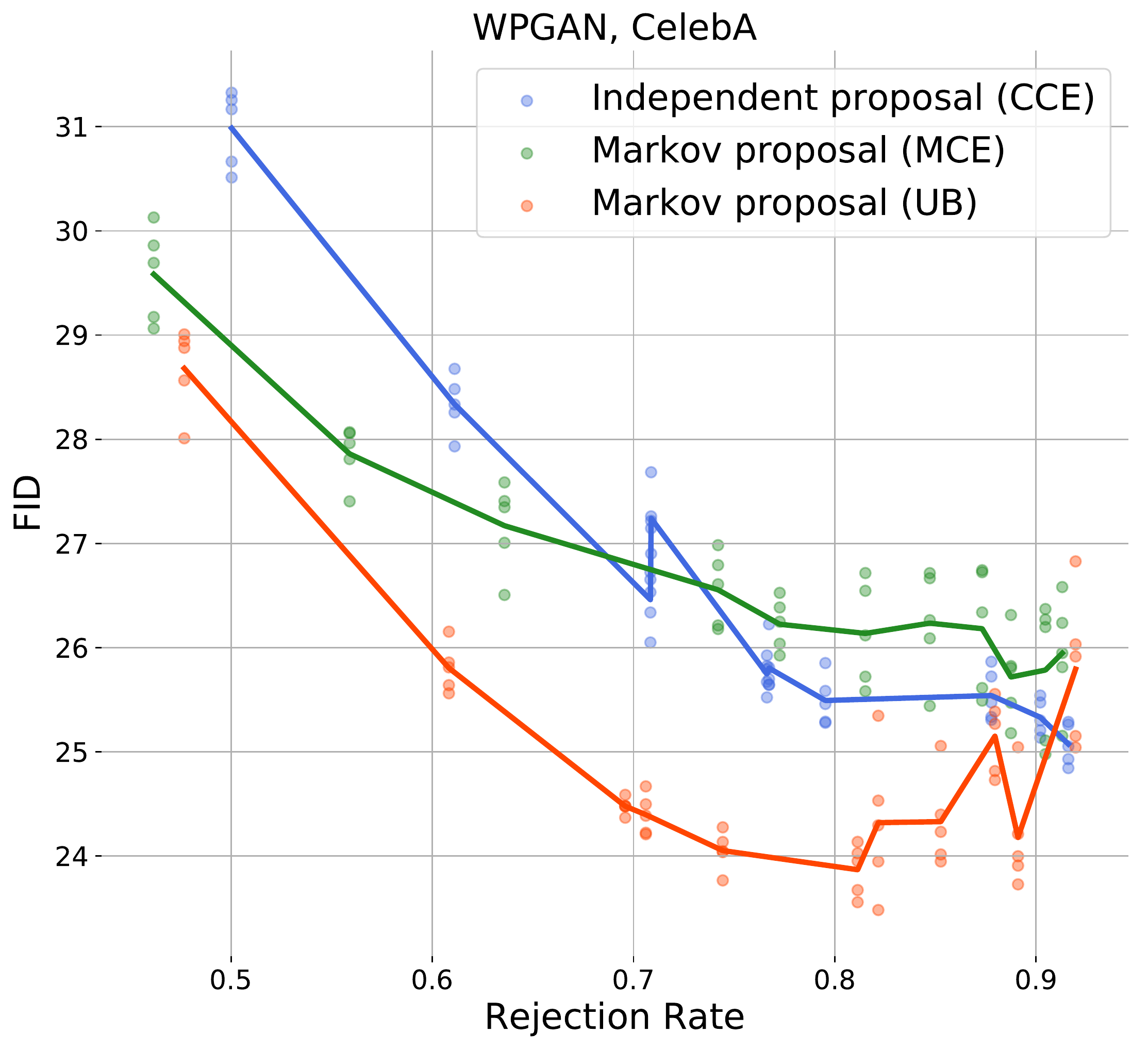}
    \includegraphics[width=0.24\textwidth]{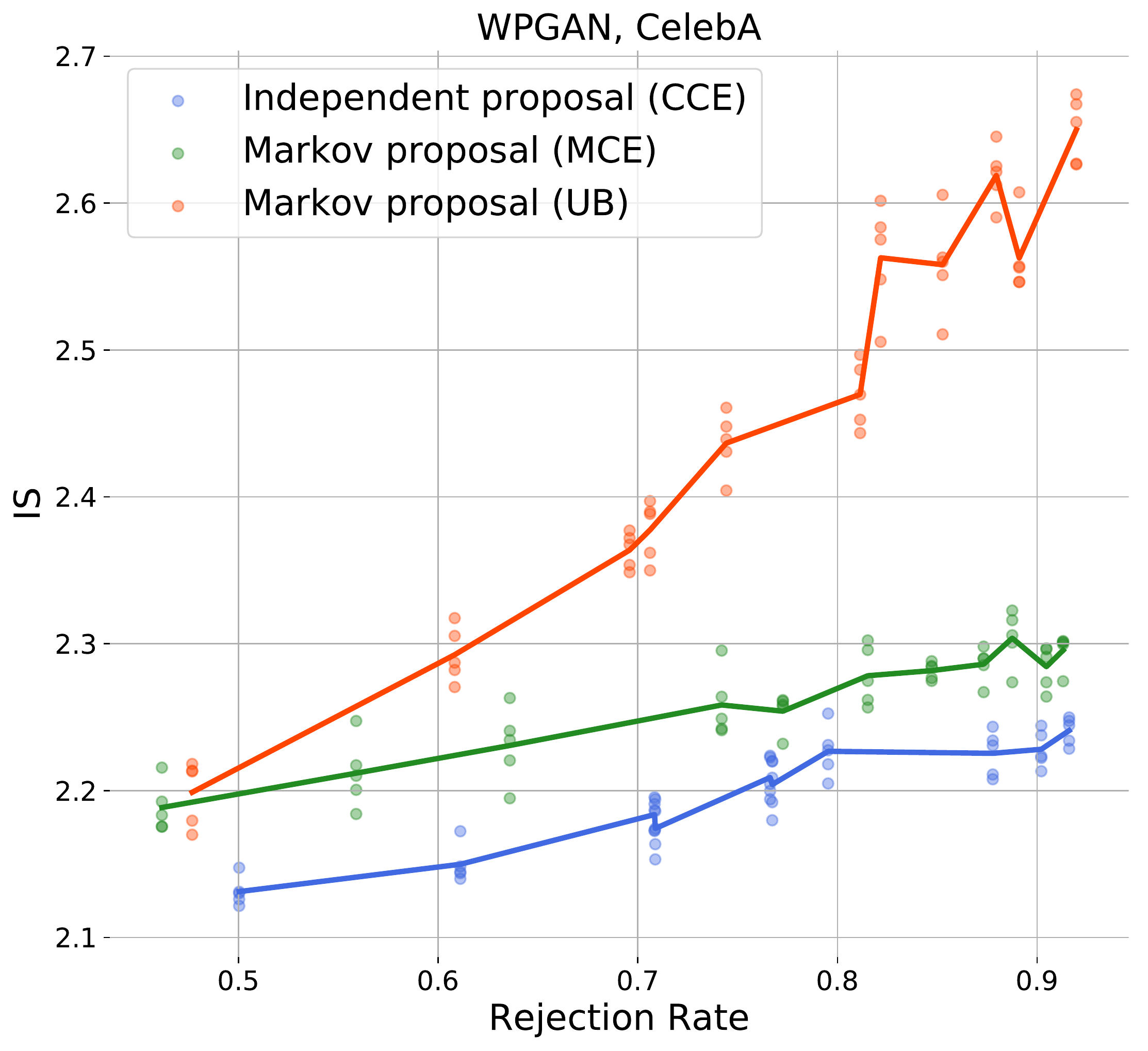}
    \caption{Comparison between different discriminators for the same generator of WPGAN in terms of performance (IS, FID) and computational efforts (rejection rate).
    Higher values of IS and lower values of FID are better.
    For a single metric evaluation, we use 10k samples.
    For every snapshot of a discriminator, we evaluate metrics $5$ times (scatter) and then average them (solid lines).}
    \label{fig:markov}
\end{figure}

In Figure \ref{fig:markov}, we demonstrate that our Markov proposal compares favorably not only against the original generator of WPGAN, but also against the chain obtained by the independent sampler (Algorithm \ref{alg:MH_disc}).
To provide the comparison, we evaluate both the performance (IS, FID) and computational efforts (rejection rate), showing that for the same rejection rate, our method results in better metrics.

\section{Conclusion}
In this paper, we propose the implicit Metropolis-Hastings algorithm for sampling from an empirical target distribution using an implicit probabilistic model as the proposal.
In the theoretical part of the paper, we upper bound the distance between the target distribution and the stationary distribution of the chain.
The contribution of the derived upper bound is two-fold.
We justify the heuristic algorithm proposed by \citep{neklyudov2018metropolis, turner2018metropolis} and derive the loss functions for the case of Markov proposal.
Moreover, the post-processing with the implicit Metropolis-Hastings algorithm can be seen as a justification or enhancement of any implicit model.
In the experimental part of the paper, we empirically validate the proposed algorithm on the real-world datasets (CIFAR-10 and CelebA).
For both tasks filtering with the proposed algorithm alleviates the gap between target and proposal distributions.

% MCMC is a well-developed field rich in theoretical results and practical applications.
% Approaching the problem of sampling from an empirical distribution from the MCMC point of view can result both in new theoretical results and novel algorithms for that problem.
% At the same time, the use of neural networks as expressive approximators in conventional MCMC algorithms can lead us to the next level of sampling techniques.
% We hope this paper will raise interest in the application of MCMC techniques to the empirical target distributions.

\bibliography{references}
\bibliographystyle{uai2019}

\newpage

\appendix

\section{The existence of stationary distribution for the transition kernel of IMH}
\label{app:prop}

Let us recall that transition kernel of the implicit Metropolis-Hastings algorithm is defined as
\begin{equation}
    t(x\cond y) =  q(x \cond y) \min\bigg\{1, \frac{d(x,y)}{d(y,x)}\bigg\} + \delta(x-y) \int dx' q(x' \cond y)\bigg(1 - \min\bigg\{1, \frac{d(x',y)}{d(y,x')}\bigg\}\bigg).
\end{equation}

In this section we show that such kernel converges to some stationary distribution if the proposal distribution $q(x\cond y)$ and the function $d(x,y)$ are \textit{continuous} and \textit{positive} on $\mathbb{R}^D\times\mathbb{R}^D$.

Firstly, we validate that such transition kernel defines a correct conditional distribution.
\begin{equation}
    t(x\cond y) \geq q(x\cond y) \min\bigg\{1, \frac{d(x,y)}{d(y,x)}\bigg\} > 0 \;\; \forall x,y \; \implies \; t_1(x) = \int t(x\cond y) t_0(y) dy > 0 \;\; \forall x
\end{equation}
Normalization constant of $t_1$ can be obtained by straightforward evaluation of the integral:
\begin{align}
    t_1(x) =& \int dy q(x \cond y)t_0(y) \min\bigg\{1, \frac{d(x,y)}{d(y,x)}\bigg\} +\\ 
    &+ \int dy \delta(x-y) t_0(y) \int dx' q(x' \cond y)\bigg(1 - \min\bigg\{1, \frac{d(x',y)}{d(y,x')}\bigg\}\bigg) \\
    t_1(x) =& \int dy q(x \cond y)t_0(y) \min\bigg\{1, \frac{d(x,y)}{d(y,x)}\bigg\} + t_0(x) - \\
    &- \int dx' q(x' \cond x)t_0(x)\min\bigg\{1, \frac{d(x',x)}{d(x,x')}\bigg\} \\
    \int t_1(x) dx =& \int dxdy q(x \cond y)t_0(y) \min\bigg\{1, \frac{d(x,y)}{d(y,x)}\bigg\} + \int t_0(x) dx -\\
    &- \int dxdx' q(x' \cond x)t_0(x)\min\bigg\{1, \frac{d(x',x)}{d(x,x')}\bigg\} \\
    \int t_1(x) dx =&  \int t_0(x) dx = 1
\end{align}

\subsection{Irreducibility}

Irreducibility of the chain can be straightforwardly proven by adaptation of the proof from 
\citep{roberts2004general}.

Consider some set $A$ such that $p(A) > 0$.
Then there exist $R > 0$ such that $p(A_R) > 0$ where $A_R = A \cap B_R(0)$ and $B_R(0)$ is a ball with radius $R$ centered at zero.
For continuous and positive $d(x,y)$ and $q(x\cond y)$ on $\mathbb{R}^D\times\mathbb{R}^D$ there exist $\eps > 0$ such that
\begin{equation}
    \inf_{x,y \in A_R} q(x\cond y) \min\bigg\{1, \frac{d(x,y)}{d(y,x)}\bigg\} \geq 
    \inf_{x,y \in B_R} q(x\cond y) \min\bigg\{1, \frac{d(x,y)}{d(y,x)}\bigg\} \geq \eps.
\end{equation}
Hence
\begin{equation}
    t(A\cond y) \geq t(A_R\cond y) \geq \int_{A_R} q(x\cond y) \min\bigg\{1, \frac{d(x,y)}{d(y,x)}\bigg\} dx \geq \eps |A_R| > 0.
\end{equation}
Thus the chain defined by $t(x\cond y)$ is irreducible.

\subsection{Aperiodicity}

Aperiodicity of the chain can be straightforwardly proven by adaptation of the proof from 
\citep{roberts2004general}.

Assume there exist two disjoint sets $A_1$ and $A_2$, such that for any starting point $y \in A_1$ the transition $t(x\cond y)$ ends in $A_2$, i.e. $t(A_2\cond y) = 1$.
However, by positivity of $d(x,y)$ and $q(x\cond y)$ we have
\begin{equation}
    t(A_1\cond y) = \int_{A_1} q(x\cond y) \min\bigg\{1, \frac{d(x,y)}{d(y,x)}\bigg\} dx > 0
    \implies t(A_2\cond y) < 1.
\end{equation}

\section{Proof of Proposition \ref{th:uniform}}
\label{app:uniform}

We consider some ergodic chain with kernel $t(x\cond y)$ and assume that $t(x \cond y)$ satisfy minorization condition, i.e. for some distribution $\nu$ and some $\eps > 0$ the following inequality holds:
\begin{equation}
    t(x\cond y) \geq \eps\nu(x), \;\; \forall x,y.
\end{equation}
We denote a distribution after $n$ steps of $t(x\cond y)$ as $t_n(x\cond y)$.
Such distribution is defined by the recurrent formula:
\begin{equation}
    t_{n+1}(x) = \int t(x\cond y) t_{n-1}(y) dy.
\end{equation}
Denoting the difference between two consequent distributions as $\Delta_n$, we study how the operator $t(x\cond y)$ changes the $l_1$-norm of $\Delta_n$.
\begin{align}
    t_{n+1}(y) = t_{n}(y) + \Delta_n(y) \implies \int t(x\cond y)t_{n+1}(y) dy = \int t(x\cond y) t_n(y) dy + \int t(x\cond y)\Delta_n(y) dy
\end{align}
Therefore
\begin{equation}
    \norm{t_{n+1} - t_n}_{TV} = \frac{1}{2} \int |\Delta_n(y)| dy, \;\; \text{and} \;\; \norm{t_{n+2} - t_{n+1}}_{TV} = \frac{1}{2} \int \bigg|\int t(x\cond y)\Delta_n(y) dy\bigg| dx.
\label{eqapp:tv_steps}
\end{equation}
Note that $\Delta_n$ integrates in zero
\begin{equation}
    \int \Delta_n(y) dy = \int t_{n+1}(y) dy - \int t_{n}(y) dy = 0.
\end{equation}
Using that fact we can rewrite the following integral
\begin{align}
    &\int t(x\cond y) \Delta_n(y) dy =  \int (t(x\cond y) - \eps \nu(x)) \Delta_n(y) dy \\
    \frac{1}{2} \int \bigg|&\int t(x\cond y)\Delta_n(y) dy\bigg| dx \leq \frac{1}{2}\int (t(x\cond y) - \eps \nu(x)) |\Delta_n(y)| dydx = (1-\eps)
    \frac{1}{2} \int |\Delta_n(y)| dy
\end{align}
Using the last inequality and equalities from \eqref{eqapp:tv_steps}, we obtain
\begin{equation}
    \norm{t_{n+2} - t_{n+1}}_{TV} \leq (1-\eps)\norm{t_{n+1} - t_{n}}_{TV}.
\end{equation}

\section{Proof of Proposition \ref{th:tv}}
\label{app:step}

For the kernel of implicit Metropolis-Hastings algorithm:
\begin{align}
    t(x\cond y) =  q(x \cond y) \min\bigg\{1, \frac{d(x,y)}{d(y,x)}\bigg\} + \delta(x-y) \int dx' q(x' \cond y)\bigg(1 - \min\bigg\{1, \frac{d(x',y)}{d(y,x')}\bigg\}\bigg),
\end{align}
we want to derive upper bound on the length of the first step in terms of TV-distance
\begin{equation}
    \norm{t_1 - p}_{TV} = \frac{1}{2} \int dx \bigg|\int dy t(x\cond y)p(y) - p(x)\bigg|.
\label{eqapp:tv_onestep}
\end{equation}
Firstly, we take the integral inside of TV-distance:
\begin{align}
    \int dy t(x\cond y) p(y) = & \int dy q(x \cond y)p(y) \min\bigg\{1, \frac{d(x,y)}{d(y,x)}\bigg\} + \int dy \delta(x-y)p(y) - \\
    & - \int dx'dy \delta(x-y)q(x' \cond y)p(y)\min\bigg\{1, \frac{d(x',y)}{d(y,x')}\bigg\} = \\
    = \int dy q(x \cond y)p(y) & \min\bigg\{1, \frac{d(x,y)}{d(y,x)}\bigg\} + p(x) -\int dx' q(x' \cond x)p(x)\min\bigg\{1, \frac{d(x',x)}{d(x,x')}\bigg\} =\\
    = \int dy q(x \cond y)p(y) & \min\bigg\{1, \frac{d(x,y)}{d(y,x)}\bigg\} + p(x) -\int dy q(y \cond x)p(x)\min\bigg\{1, \frac{d(y,x)}{d(x,y)}\bigg\}
\end{align}
Substituting this formula into \eqref{eqapp:tv_onestep} we obtain
\begin{align}
    \norm{t_1 - p}_{TV}  = &  \frac{1}{2} \int dx \bigg| \int dy q(x \cond y)p(y) \min\bigg\{1, \frac{d(x,y)}{d(y,x)}\bigg\} -  \int dy q(y \cond x)p(x)\min\bigg\{1, \frac{d(y,x)}{d(x,y)}\bigg\}\bigg| \leq \\
    \leq & \frac{1}{2} \int dxdy \bigg|q(x \cond y)p(y) \min\bigg\{1, \frac{d(x,y)}{d(y,x)}\bigg\} -  q(y \cond x)p(x)\min\bigg\{1, \frac{d(y,x)}{d(x,y)}\bigg\}\bigg| = 
\end{align}
Note that changing variables in integral does not change value of function, hence we can integrate over the half of the space and then multiply the integral by $2$:
\begin{align}
    = &\int_{A} dxdy \bigg|q(x \cond y)p(y) \min\bigg\{1, \frac{d(x,y)}{d(y,x)}\bigg\} -  q(y \cond x)p(x)\min\bigg\{1, \frac{d(y,x)}{d(x,y)}\bigg\}\bigg| = \\ 
    & A = \bigg\{x,y : \frac{d(x,y)}{d(y,x)} \geq 1 \bigg\} \\
    = &\int_{A} dxdy \bigg|q(x \cond y)p(y) -  q(y \cond x)p(x)\min\bigg\{1, \frac{d(y,x)}{d(x,y)}\bigg\}\bigg|
\end{align}
Thus, we obtain 
\begin{equation}
    \norm{t_1 - p}_{TV} \leq 2 \norm{q(x \cond y)p(y) -  q(y \cond x)p(x)\min\bigg\{1, \frac{d(y,x)}{d(x,y)}\bigg\}}_{TV}
\end{equation}

\section{Proof of Proposition \ref{th:pinsker}}
\label{app:pinsker}

To prove Proposition \ref{th:pinsker} we extend the proof from \citep{pollard2000asymptopia}.
Consider a distribution $\alpha(x)$ and some positive function $f(x) > 0 \;\; \forall x$.
Normalization constants for $\alpha$ and $f$ are
\begin{equation}
    \int \alpha(x) dx = 1, \;\;\; \text{ and } \;\;\; \int f(x) dx = C.
\end{equation}
The proof is constructed around the following inequality
\begin{equation}
    (1+r)\log(1+r) -r \geq \frac{1}{2}\frac{r^2}{1+r/3}, \;\; r \geq -1.
\label{eqapp:ineq}
\end{equation}
For $r$ we consider the ratio $r(x) = \alpha(x)/f(x) - 1$, and introduce a random variable $F$ with the density $f(x)/C$.
Then 
\begin{align}
    &\mean_F r(x) = \int \frac{f(x)}{C}\bigg(\frac{\alpha(x)}{f(x)}-1\bigg)dx = \frac{1}{C} - 1 \\
    &\mean_F (1+r(x)) \log(1+r(x)) = \frac{1}{C}\int \alpha(x) \log  \frac{\alpha(x)}{f(x)} \triangleq \frac{1}{C}\widehat{\mathrm{KL}}(\alpha\Vert f)\\
    &\mean_F \bigg(1 +\frac{r(x)}{3} \bigg) = \frac{2}{3} + \frac{1}{3C} > 0 \\
    &\mean_F |r(x)| = \frac{1}{C}\int \bigg|\alpha(x)-f(x) \bigg|dx = \frac{2}{C}\norm{\alpha - f}_{TV}
\end{align}
Substituting all the equations into \eqref{eqapp:ineq} we obtain
\begin{align}
    \mean_F \bigg[(1+r(x)) \log(1+r(x)) - r(x) \bigg] &\geq \frac{1}{2} \mean_F \bigg[\frac{r(x)^2}{1+r(x)/3}\bigg] \\
    \mean_F \bigg(1 +\frac{r(x)}{3} \bigg) \mean_F \bigg[(1+r(x)) \log(1+r(x)) - r(x) \bigg] &\geq \frac{1}{2} \mean_F \bigg[\frac{r(x)^2}{1+r(x)/3}\bigg] \mean_F \bigg(1 +\frac{r(x)}{3} \bigg)\\
    \mean_F \bigg(1 +\frac{r(x)}{3} \bigg) \mean_F \bigg[(1+r(x)) \log(1+r(x)) - r(x) \bigg] &\geq \frac{1}{2}  \bigg[\mean_F|r(x)|\bigg]^2\\
    \frac{2C + 1}{3C}\bigg(\frac{1}{C}\widehat{\mathrm{KL}}(\alpha\Vert f) - \frac{1}{C} + 1\bigg) &\geq \frac{2}{C^2}\norm{\alpha - f}_{TV}^2
\end{align}
Hence, we obtain
\begin{equation}
    \norm{\alpha - f}_{TV}^2 \leq \frac{2C + 1}{6}\bigg(\widehat{\mathrm{KL}}(\alpha\Vert f) + C - 1\bigg)
\end{equation}
Note, that if $f$ is a distribution, then $C = 1$ and we obtain Pinsker's inequality:
\begin{equation}
    \norm{\alpha - f}_{TV}^2 \leq \frac{1}{2}\widehat{\mathrm{KL}}(\alpha\Vert f).
\end{equation}

\section{DRE}
\label{app:dre}

We derive the formula for the optimal discriminator by taking derivative of the following objective w.r.t. the value of $d(x,y)$ in a single point $(x,y)$
\begin{equation}
    \min_d \text{\Large $\mean$}_{\text{\footnotesize $\begin{matrix}
    x \sim p(x) \\ 
    y \sim q(y\cond x)
    \end{matrix}$}} \bigg[ \log\frac{d(y,x)}{d(x,y)} + \frac{d(y,x)}{d(x,y)} \bigg].
\end{equation}
Speaking informally, we treat the expectation as a sum over all the possible points.
Taking a derivative w.r.t. a single point allows us to consider only two elements of the sum.
\begin{align}
    \nabla_{d(x,y)} \bigg( &p(x)q(y\cond x)  \bigg[ \log\frac{d(y,x)}{d(x,y)} + \frac{d(y,x)}{d(x,y)} \bigg] + p(y)q(x\cond y) \bigg[ \log\frac{d(x,y)}{d(y,x)} + \frac{d(x,y)}{d(y,x)} \bigg] \bigg) = 0\\
    &p(x)q(y\cond x)\bigg[-\frac{1}{d(x,y)} - \frac{d(y,x)}{d(x,y)^2}\bigg] + p(y)q(x\cond y) \bigg[ \frac{1}{d(x,y)} + \frac{1}{d(y,x)} \bigg] = 0 \\
    &\frac{p(x)q(y\cond x)}{p(y)q(x\cond y)}\bigg[-1 - \frac{d(y,x)}{d(x,y)}\bigg] + \bigg[ 1 + \frac{d(x,y)}{d(y,x)} \bigg] = 0 \\
    &\frac{p(x)q(y\cond x)}{p(y)q(x\cond y)}\frac{d(y,x) + d(x,y)}{d(x,y)} = \frac{d(x,y)+d(y,x)}{d(y,x)} \\
    &\frac{p(x)q(y\cond x)}{p(y)q(x\cond y)} = \frac{d(x,y)}{d(y,x)}
\end{align}
Note that we do not derive an explicit form of $d(x,y)$, actually $d(x,y)$ could be any function which ratio equals to the density-ratio. 

The same result can be obtained by taking a derivative in function space, but for simplicity, we provide here an informal proof by taking the pointwise derivative.

\section{Relation to the cross-entropy}
\label{app:ce}

In Section \ref{sec:exp} we show that the obtained loss \eqref{eq:main_loss} is hard for optimization via the stochastic gradient descent.
However, in this Section we make a connection between loss \eqref{eq:main_loss} and the conventional loss for a density-ratio estimation --- cross-entropy.

\subsection{Markov proposal}

For Markov proposal, the loss from \eqref{eq:main_loss} can be straightforwardly upper bounded by the cross-entropy:
\begin{align}
    \text{\Large $\mean$}_{\text{\footnotesize $\begin{matrix}
    x \sim p(x) \\ 
    y \sim q(y\cond x)
    \end{matrix}$}} \bigg[ \log\frac{d(y,x)}{d(x,y)} + \frac{d(y,x)}{d(x,y)} \bigg] \leq \text{\Large $\mean$}_{\text{\footnotesize $\begin{matrix}
    x \sim p(x) \\ 
    y \sim q(y\cond x)
    \end{matrix}$}} \bigg[ -\log d(x,y) - \log (1-d(y,x)) + \frac{1}{b} \bigg].
\end{align}
That yields the optimal discriminator
\begin{equation}
    d(x,y) = \frac{p(x)q(y\cond x)}{p(x)q(y\cond x)+p(y)q(x\cond y)},
\end{equation}
using which we can achieve $\norm{t_{\infty} - p}_{TV} = 0$.

\subsection{Independent proposal}

In Section \ref{sec:pre} we describe Algorithm \ref{alg:MH_disc} proposed in \citep{neklyudov2018metropolis, turner2018metropolis}.
The idea of the algorithm is to use learned generator of any GAN model as \textit{independent} proposal $q(x)$ in the Metropolis-Hastings algorithm.
Authors propose to learn a discriminator $d(x)$ by minimization of the cross-entropy:
\begin{align}
    \min_d \bigg[-\mean_{x\sim p(x)} \log d(x) -\mean_{x \sim q(x)} \log (1-d(x)) \bigg],
\label{eq:ce_disc}
\end{align}
and then to estimate the density-ratio as
\begin{align}
    \frac{p(x)q(y)}{p(y)q(x)} \approx \frac{d(x)(1-d(y))}{(1-d(x))d(y)}.
\end{align}
In this section, we show that there exists such an upper bound on $\norm{t_\infty - p}_{TV}$ that its optimization is equivalent to the optimization of cross-entropy \eqref{eq:ce_disc}.
To derive such upper bound we upper bound the discriminator objective \eqref{eq:main_loss}, considering an independent proposal $q(x)$ and factorized discriminator $d(x,y) = d(x)(1-d(y))$.
\begin{align}
    \text{\Large $\mean$}_{\text{\footnotesize $\begin{matrix}
    x \sim p(x) \\ 
    y \sim q(y)
    \end{matrix}$}} \bigg[ \log\frac{d(y)(1-d(x))}{d(x)(1-d(y))} + \frac{d(y)(1-d(x))}{d(x)(1-d(y))} \bigg] \leq \text{\Large $\mean$}_{\text{\footnotesize $\begin{matrix}
    x \sim p(x) \\ 
    y \sim q(y)
    \end{matrix}$}} \bigg[ \log\frac{d(y)(1-d(x))}{d(x)(1-d(y))} + \frac{1}{b}\bigg]
\end{align}
Splitting the logarithm into sum results in
\begin{align}
\begin{split}
    \bigg[ & -\mean_{x\sim p(x)} \log d(x) -\mean_{y\sim q(y)} \log (1-d(y)) + \mean_{x\sim p(x)} \log (1-d(x)) +\mean_{y\sim q(y)} \log d(y)\bigg] \leq \\
    \leq & -\mean_{x\sim p(x)} \log d(x) -\mean_{y\sim q(y)} \log (1-d(y)),
\end{split}
\end{align}
where the last upper bound is the cross-entropy \eqref{eq:ce_disc}.
The obtained upper bound on the discriminator objective \eqref{eq:main_loss} can be substituted in \eqref{eq:final_bound} that results in
\begin{align}
\begin{split}
    &\norm{t_{\infty} - p}_{TV}^2 \leq \mathcal{L}(d) \leq \bigg(\frac{4+2b}{3\eps^2 b^3}\bigg)\cdot\\
    \cdot \bigg(& -\mean_{x\sim p(x)} \log d(x) -\mean_{y\sim q(y)} \log (1-d(y)) + \frac{1}{b} - 1 + \mathrm{KL}(q(y)p(x)\Vert q(x)p(y))\bigg).
\end{split}
\label{eq:ce_final_bound}
\end{align}
Hence, minimization of the cross-entropy leads to the minimization of the TV-distance between stationary distribution of the chain $t_\infty(x)$ and target distribution $p(x)$.
Note that during optimization of such upper-bound we also could achieve $\norm{t_{\infty} - p}_{TV} = 0$ for any target $p(x)$ and proposal $q(x)$, since the optimal discriminator $d^*(x)$ allows correct estimation of density ratio:
\begin{align}
    \frac{d^*(x)(1-d^*(y))}{(1-d^*(x))d^*(y)} = \frac{p(x)q(y)}{p(y)q(x)}.
\end{align}

\section{Synthetic example}
\label{app:synth}

We validate the proposed algorithm and compare different losses on a synthetic target distribution.
For the target empirical distribution we take $5000$ samples from the mixture of two Gaussians $p(x) = 0.5\Normal(x\cond \mu=-2,\sigma=0.5) + 0.5\Normal(x\cond \mu=2,\sigma=0.7)$.
We imitate an implicit Markov proposal by sampling from the random-walk kernel $q(x\cond y)=\Normal(x\cond \mu=y,\sigma=1.0)$, and an implicit independent proposal by sampling from the Gaussian $q(x)=\Normal(x\cond \mu=0.0,\sigma=2.0)$.
Note, despite that we know densities of the target and proposals, we use only samples from these distributions during training and sampling stages.
As a discriminator, we use the neural network with 3 fully-connected layers ($100$ hidden neurons) and learn it with the Adam optimizer for $1000$ iterations.

Since we have access to the density of distributions, we use the TV-distance from
\eqref{eq:tv_bound} as a test metric.
Such a metric can be treated as averaged $l_1$ error of the density-ratio estimation error:
\begin{align}
    2\norm{q(y\cond x)p(x) - q(x \cond y)p(y) \frac{d(x,y)}{d(y,x)}}_{TV} = \int dxdy~q(x \cond y)p(y) \bigg|\frac{q(y\cond x)p(x)}{q(x \cond y)p(y)} - \frac{d(x,y)}{d(y,x)}\bigg|.
\label{eq:test_tv}
\end{align}

We compare losses from Table \ref{tab:losses} in Figure \ref{fig:tv}.
For Markov proposal (left plot in Fig. \ref{fig:tv}), the optimization of upper bound (UB) behaves similarly to the optimization of cross-entropy (MCE).
However, for the independent proposal (right plot in Fig. \ref{fig:tv}), the best metric for optimization is the conventional cross-entropy (CCE).
In Figure \ref{fig:synth_example}, we demonstrate filtering of the independent proposal with the discriminator learned by the optimization of cross-entropy (CCE).

Note that learning a discriminator for the random-walk proposal allows for estimation of target unnormalized density:
\begin{equation}
    \frac{d(x,y)}{d(y,x)} \approx \frac{p(x)q(y\cond x)}{p(y)q(x\cond y)} = \frac{p(x)}{p(y)},
\label{eq:qeq}
\end{equation}
since $q(x\cond y) = q(y\cond x)$.

\begin{minipage}[h]{0.48\textwidth}
\centering
    \includegraphics[width=0.49\textwidth]{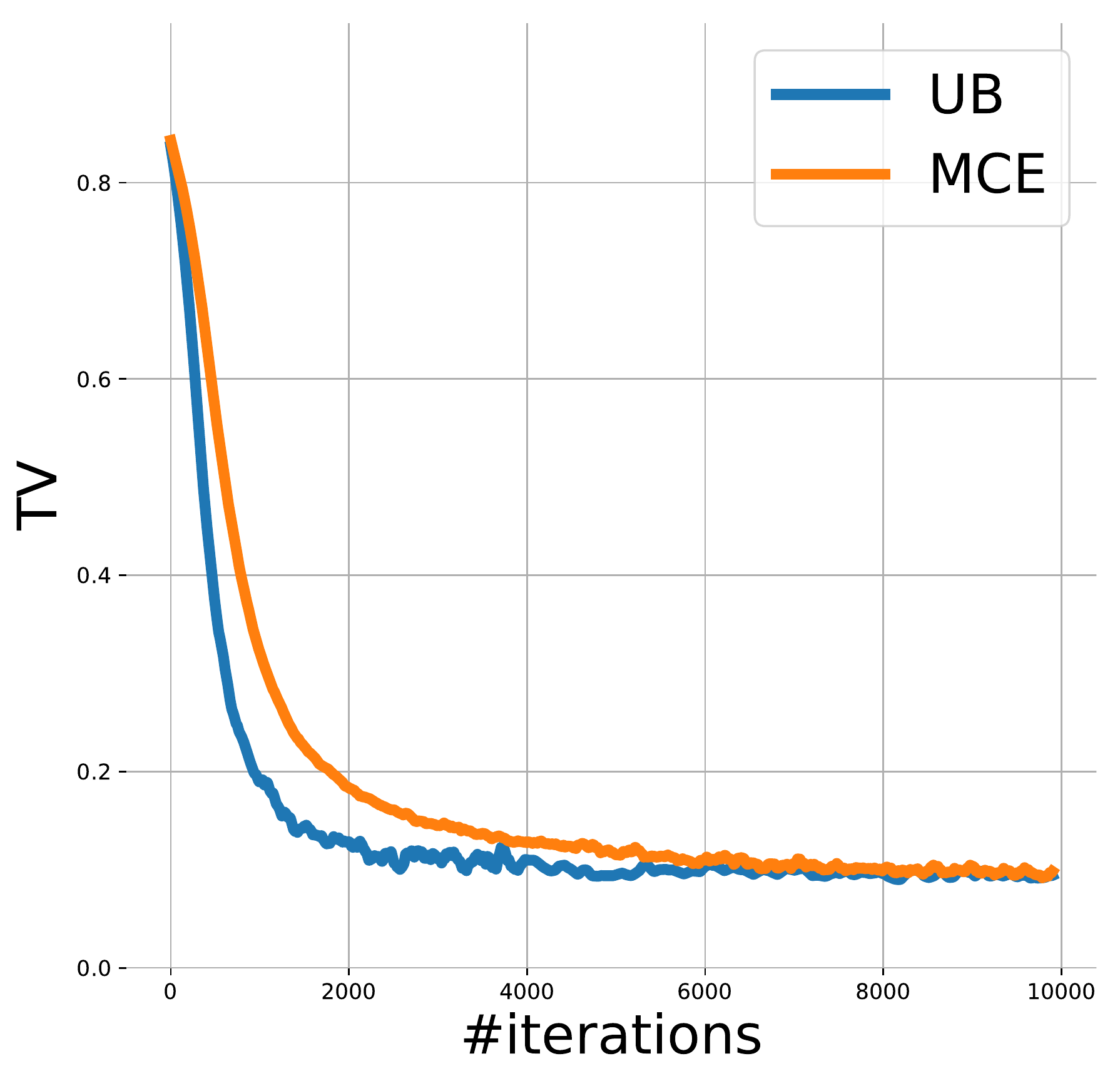}
    \includegraphics[width=0.49\textwidth]{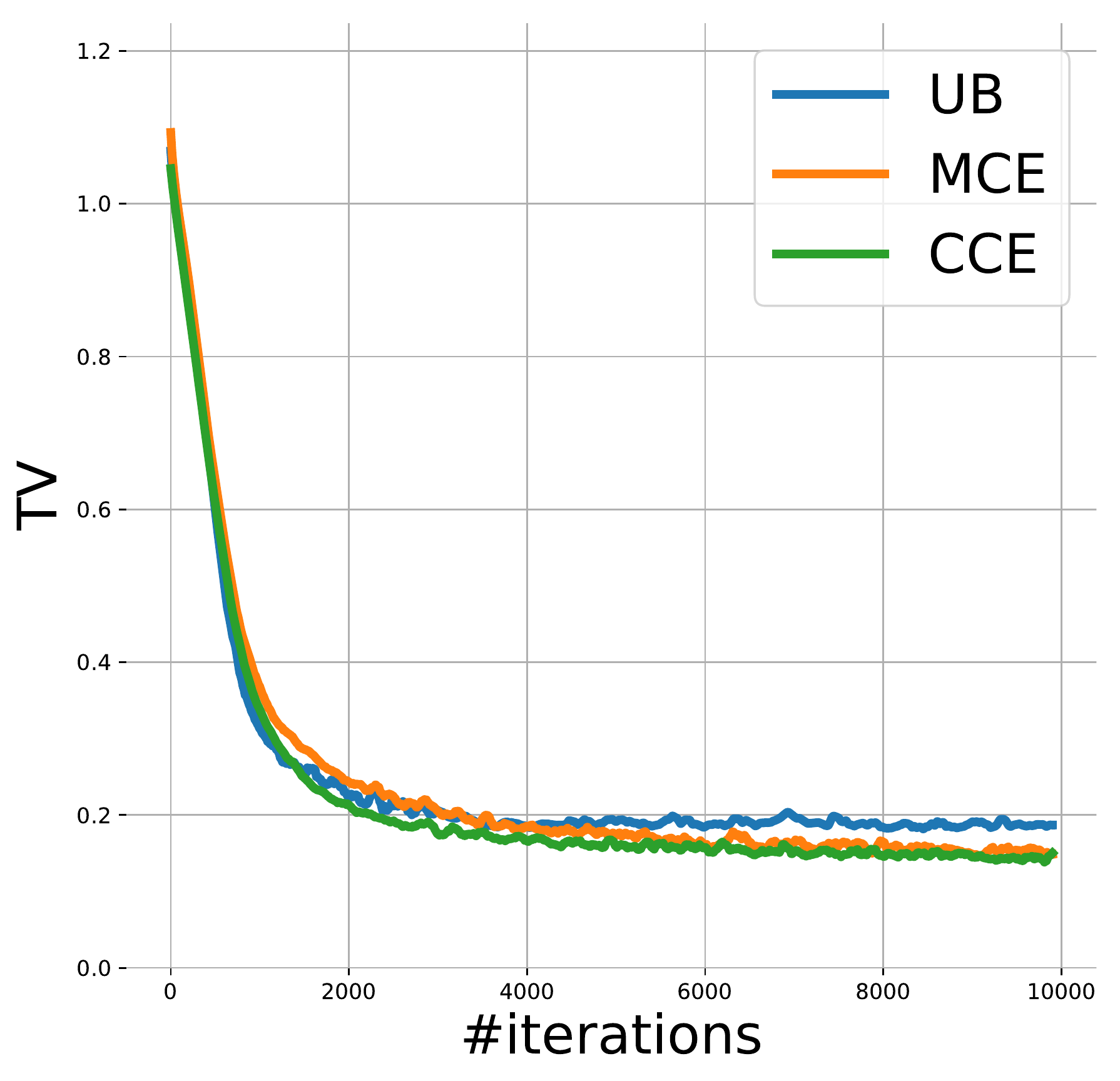}
    \captionof{figure}{
    Comparison of different losses for a discriminator in terms of the TV-distance \eqref{eq:test_tv}.
    On the left plot we learn the discriminator for the Markov proposal, on the right plot we learn the discriminator for the independent proposal.
    For losses see Table \ref{tab:losses}.}
    \label{fig:tv}
\end{minipage}\hfill
\begin{minipage}[h]{0.48\textwidth}
\centering
    \includegraphics[width=\textwidth]{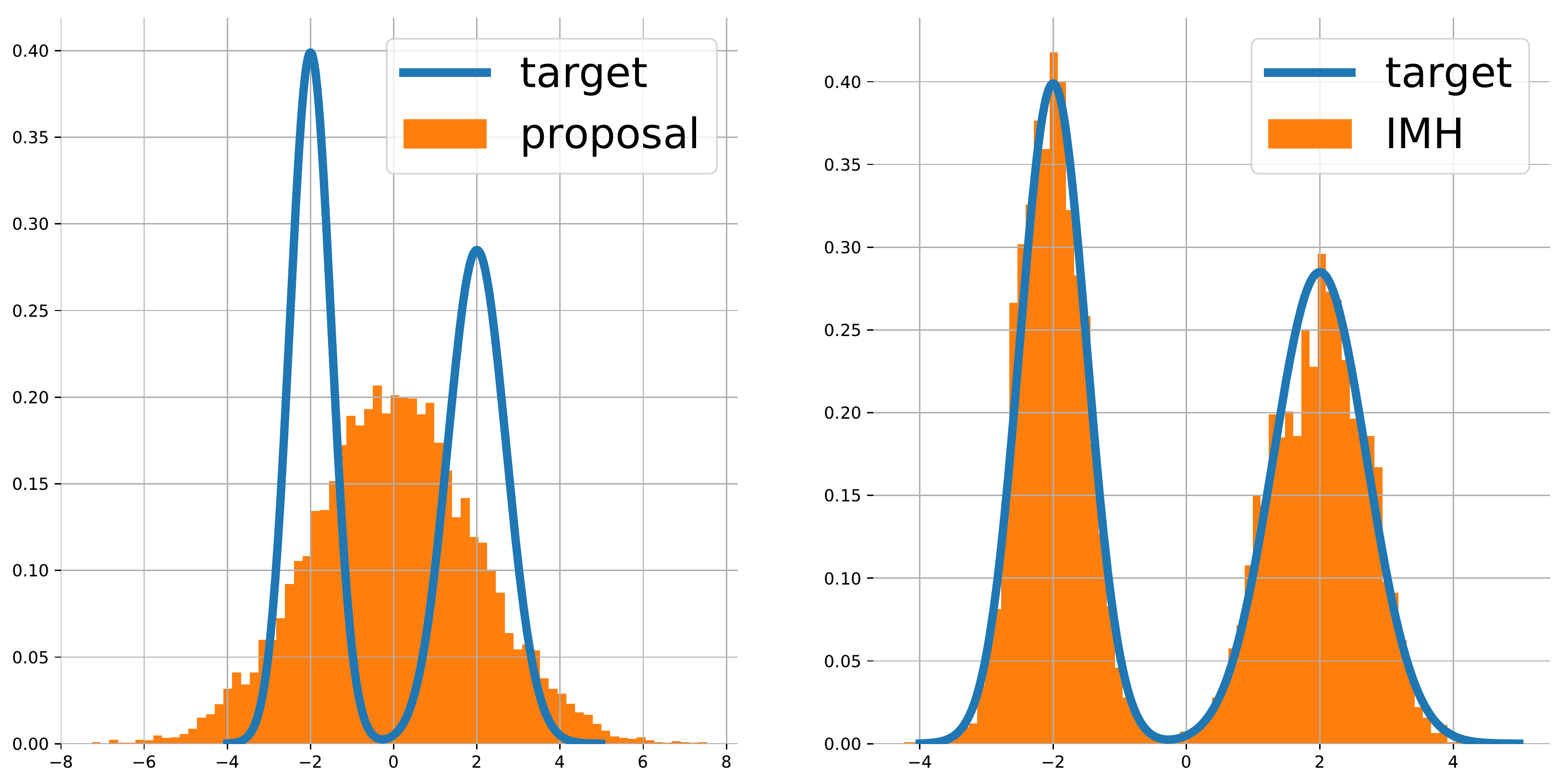}
    \captionof{figure}{
    Samples from the independent proposal distribution are on the left.
    Samples obtained after filtering with the implicit Metropolis-Hastings (IMH) algorithm are on the right.}
    \label{fig:synth_example}
\end{minipage}

\section{Monotonous improvements throught the learning of discriminator}
\label{app:mono_plots}
\begin{figure}[h]
    \centering
    \includegraphics[width=0.49\textwidth]{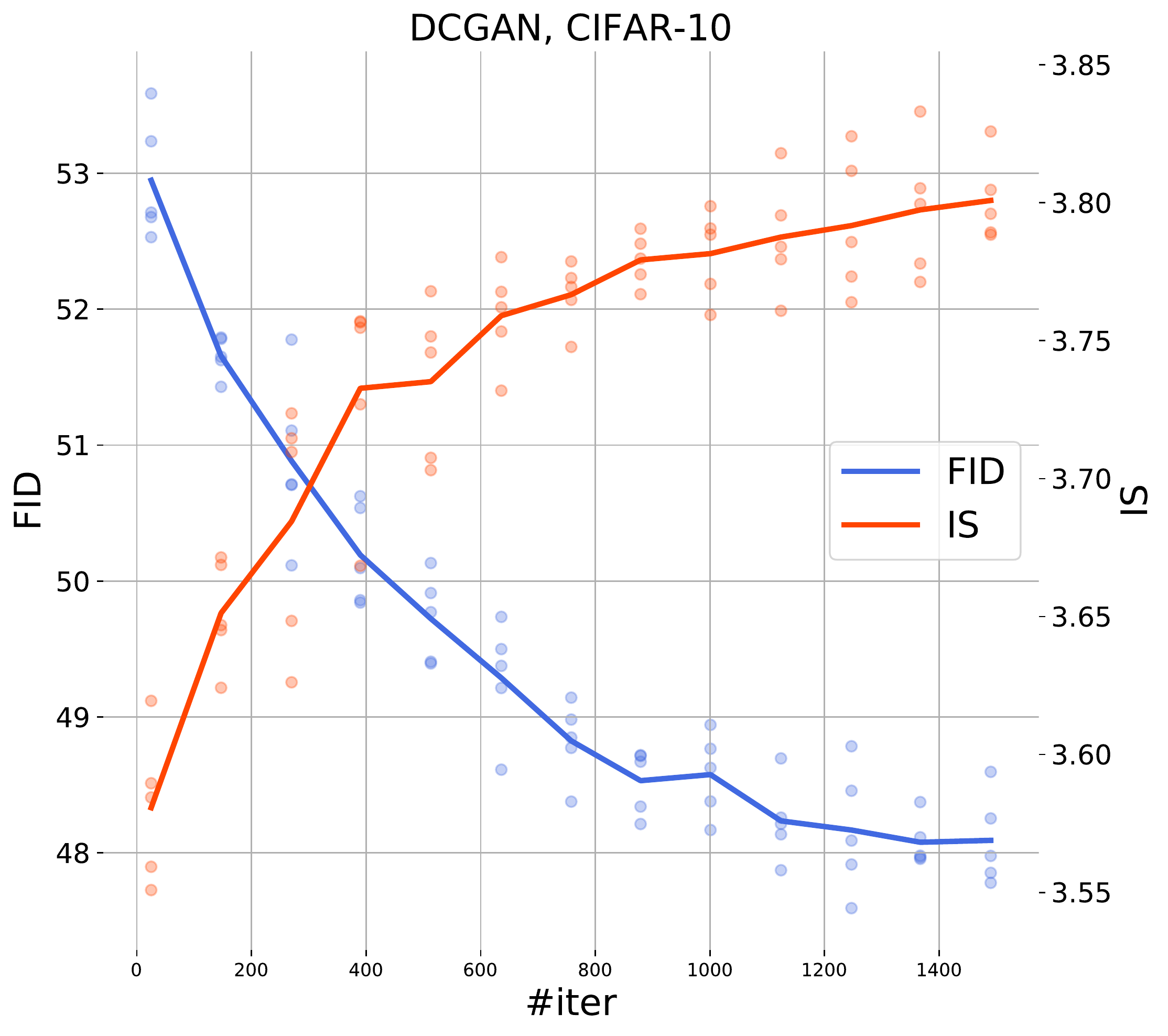}
    \includegraphics[width=0.49\textwidth]{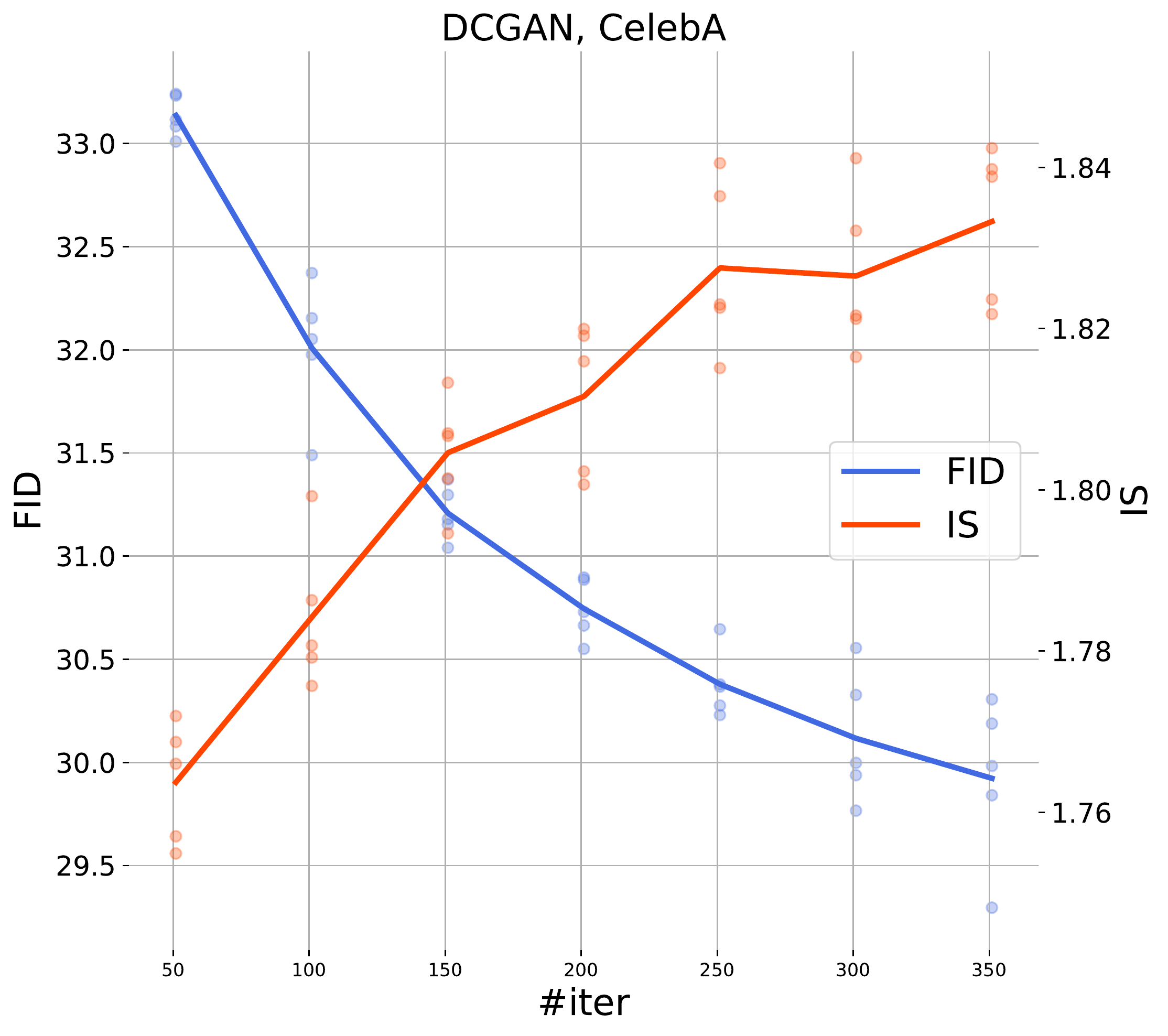}
    \caption{Monotonous improvements in terms of FID and IS for the learning of discriminator by CCE.
    During iterations, we evaluate metrics $5$ times (scatter) and then average them (solid lines).
    For a single metric evaluation, we use 10k samples. 
    Higher values of IS and lower values of FID are better.
    Performance for the original generator corresponds to $0$th iteration of a discriminator.}
    \label{fig:iters}
\end{figure}

\begin{figure}[h]
    \centering
    \includegraphics[width=0.49\textwidth]{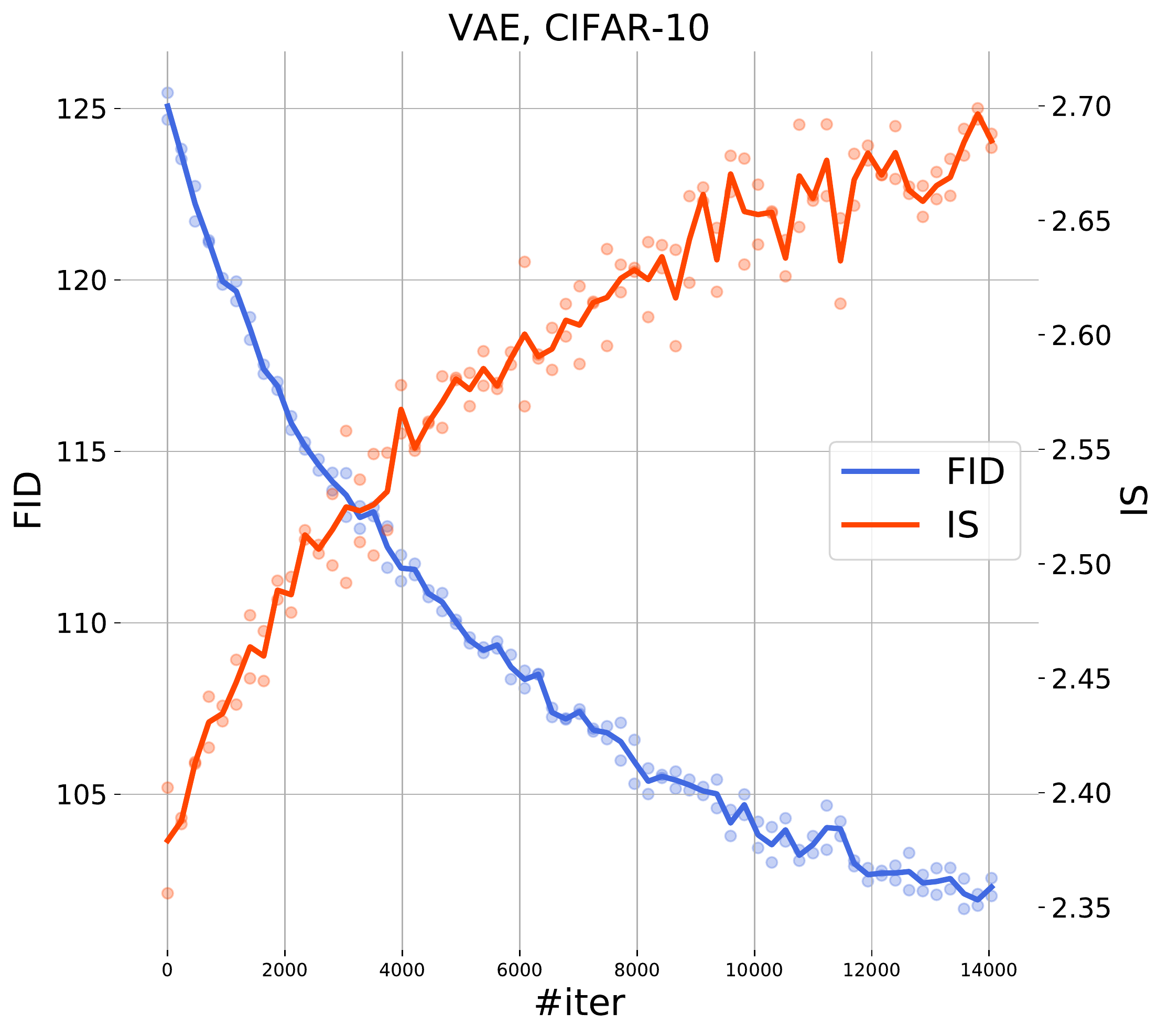}
    \includegraphics[width=0.49\textwidth]{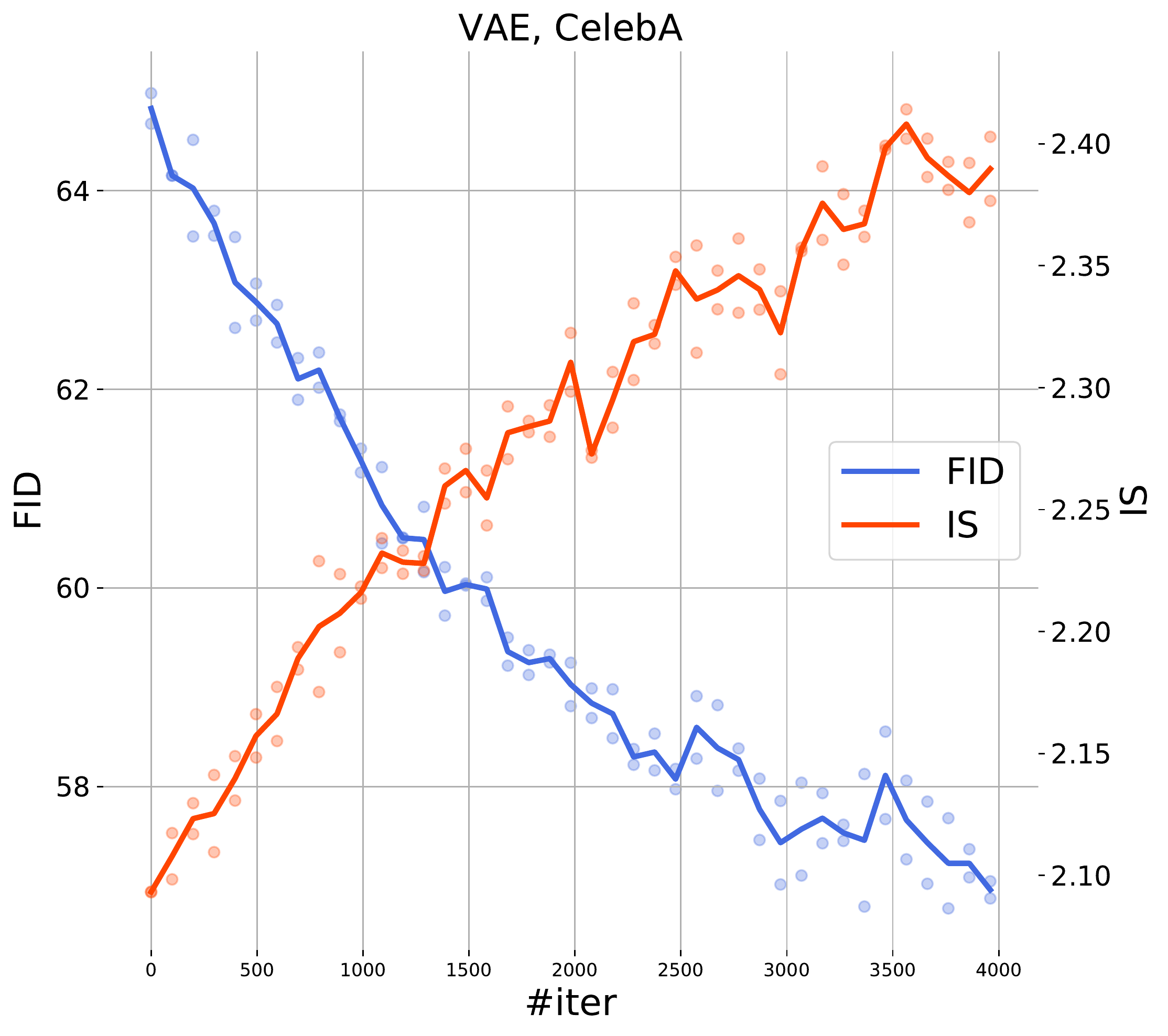}\\
    \includegraphics[width=0.49\textwidth]{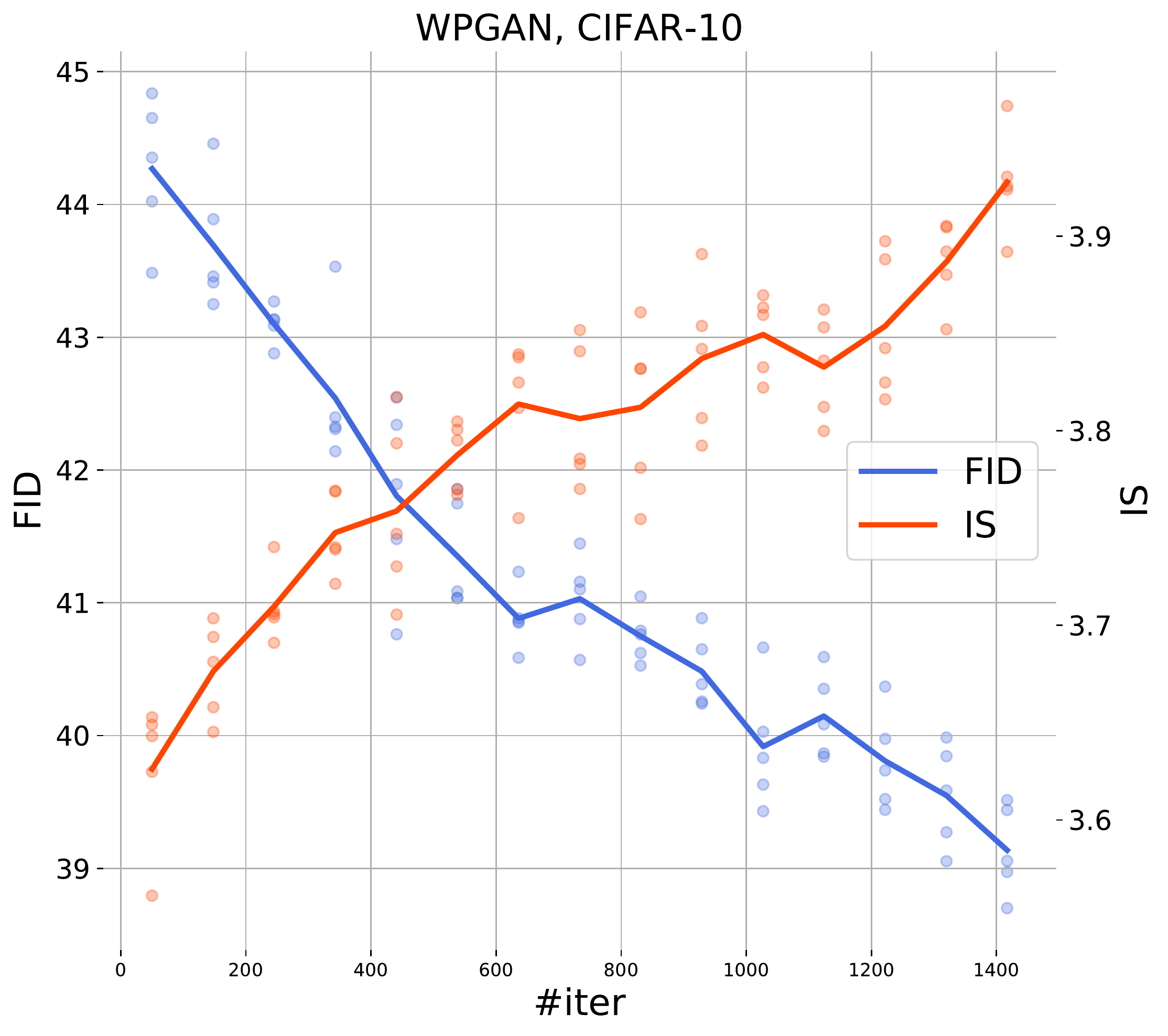}
    \includegraphics[width=0.49\textwidth]{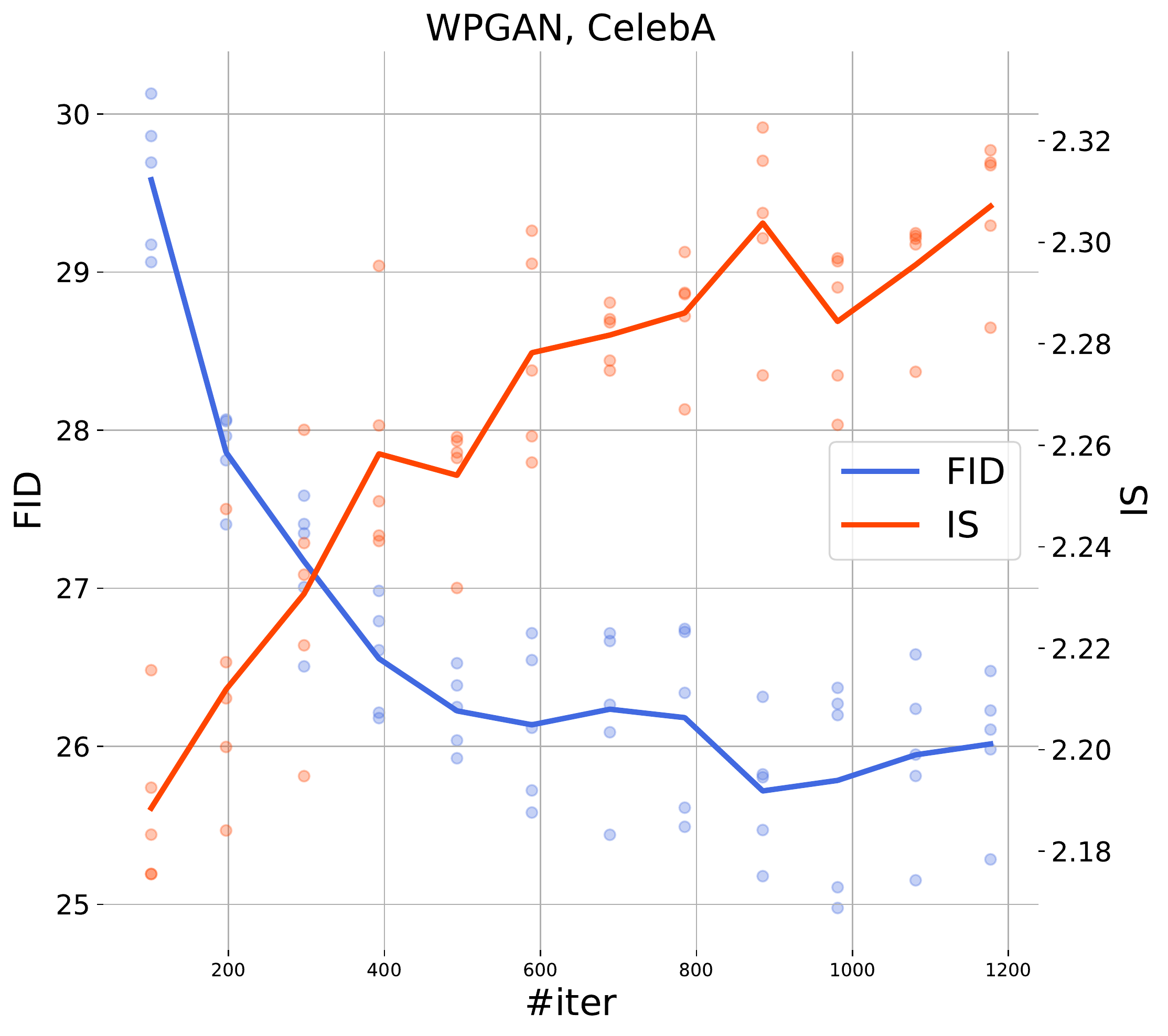}
    \caption{Monotonous improvements in terms of FID and IS for the learning of discriminator by CCE.
    During iterations, we evaluate metrics $5$ times (scatter) and then average them (solid lines).
    For a single metric evaluation, we use 10k samples. 
    Higher values of IS and lower values of FID are better.
    Performance for the original generator corresponds to $0$th iteration of a discriminator.}
\end{figure}
\newpage

\section{Losses for DRE}
\label{app:losses}
\begin{table*}[t]
\def\arraystretch{3.5}
\caption{Different losses for the density-ratio estimation.}
\label{tab:all_losses}
\begin{center}
\vskip 0.15in
\begin{small}
% \begin{sc}
\begin{adjustbox}{angle=90}
\resizebox{\textheight}{!}{
\begin{tabular}{l|l|l|l}
\toprule
{\bf PROPOSAL} & {\bf NAME} & {\bf LOSS} &{\bf DRE} \\
\midrule
\multirow{3}{*}{MARKOV} & UPPER BOUND (UB) & $\displaystyle\int dxdy~p(x)q(y\cond x)\bigg[ \log\frac{d(y,x)}{d(x,y)} + \frac{d(y,x)}{d(x,y)}\bigg]$ & $\displaystyle\frac{p(x)q(y\cond x)}{p(y)q(x\cond y)}=\frac{d(x,y)}{d(y,x)}$ \\
    & MARKOV CROSS-ENTROPY (MCE) & $\displaystyle\int dxdy~p(x)q(y\cond x)[-\log d(x,y) -\log (1-d(y,x))]$ 
    & $\displaystyle\frac{p(x)q(y\cond x)}{p(y)q(x\cond y)}=\frac{d(x,y)}{d(y,x)}$ \\
    & LINEAR TERM (LT) & $\displaystyle\int dxdy~p(x)q(y\cond x)\bigg[\frac{d(y,x)}{d(x,y)}\bigg]$ 
    & $\displaystyle\frac{p(x)q(y\cond x)}{p(y)q(x\cond y)}=\bigg(\frac{d(x,y)}{d(y,x)}\bigg)^2$ \\ \hline
\multirow{5}{*}{INDEPENDENT} & UPPER BOUND (UB) & $\displaystyle\int dxdy~p(x)q(y)\bigg[ \log\frac{d(y)(1-d(x))}{d(x)(1-d(y))} + \frac{d(y)(1-d(x))}{d(x)(1-d(y))}\bigg]$ & $\displaystyle\frac{p(x)q(y)}{p(y)q(x)}=\frac{d(x)(1-d(y))}{d(y)(1-d(x))}$ \\
    & \multirow{2}{*}{MARKOV CROSS-ENTROPY (MCE)} & \multirow{2}{*}{$\displaystyle\int dxdy~p(x)q(y)[-\log d(x)(1-d(y)) -\log (1-d(y)(1-d(x)))]$} 
    & $\displaystyle\frac{p(x)q(y)}{p(y)q(x)}=\frac{d(x)(1-d(y)(1-d(x)))}{(1-d(x))(1-d(x)(1-d(y)))}\cdot$ \\
    & & & $\;\;\;\;\;\;\;\;\;\;\;\;\;\;\displaystyle\cdot\frac{(2(1-d(x)(1-d(y)))-d(y))}{((1-d(y)(1-d(x)))+d(y)d(x))}$ \\
    & LINEAR TERM (LT) & $\displaystyle\int dxdy~ p(x)q(y)\bigg[\frac{d(y)(1-d(x))}{d(x)(1-d(y))}\bigg]$ 
    & $\displaystyle\frac{p(x)q(y)}{p(y)q(x)}=\bigg(\frac{d(x)(1-d(y))}{d(y)(1-d(x))}\bigg)^2$ \\
    & CONVENTIONAL CROSS-ENTROPY (CCE) & $\displaystyle\int dxdy~p(x)q(y)[-\log d(x)(1-d(y))]$ 
    & $\displaystyle\frac{p(x)q(y)}{p(y)q(x)}=\frac{d(x)(1-d(y))}{d(y)(1-d(x))}$ \\
\bottomrule
\end{tabular}}
\end{adjustbox}
% \end{sc}
\end{small}
\vskip -0.1in
\end{center}
\end{table*}

\end{document}